\documentclass[review,authoryear]{elsarticle}
\journal{Expert Systems With Applications}
\usepackage{cite}
\usepackage{graphicx}
\usepackage{amsmath}
\usepackage[caption=false,font=footnotesize]{subfig}
\usepackage{url}
\usepackage[strings]{underscore} 
\usepackage{amssymb}  
\usepackage{ctable}
\usepackage[ruled,vlined,linesnumbered]{algorithm2e}
\begin{document}

\title{Interactive Image Segmentation using Label Propagation through Complex Networks}

\author{Fabricio Breve}
\ead{fabricio.breve@unesp.br}

\address{Institute of Geosciences and Exact Sciences,
S\~{a}o Paulo State University (UNESP), Rio Claro, SP, 13506-900, Brazil}

\begin{abstract}
Interactive image segmentation is a topic of many studies in image processing. In a conventional approach, a user marks some pixels of the object(s) of interest and background, and an algorithm propagates these labels to the rest of the image. This paper presents a new graph-based method for interactive segmentation with two stages. In the first stage, nodes representing pixels are connected to their $k$-nearest neighbors to build a complex network with the small-world property to propagate the labels quickly. In the second stage, a regular network in a grid format is used to refine the segmentation on the object borders. Despite its simplicity, the proposed method can perform the task with high accuracy. Computer simulations are performed using some real-world images to show its effectiveness in both two-classes and multi-classes problems. It is also applied to all the images from the Microsoft GrabCut dataset for comparison, and the segmentation accuracy is comparable to those achieved by some state-of-the-art methods, while it is faster than them. In particular, it outperforms some recent approaches when the user input is composed only by a few ``scribbles'' draw over the objects. Its computational complexity is only linear on the image size at the best-case scenario and linearithmic in the worst case.
\end{abstract}


\maketitle

\section{Introduction}

Image segmentation is the process of dividing an image in parts, identifying objects or other relevant information \citep{Shapiro2001}. It is one of the most difficult tasks in image processing \citep{Gonzalez2008}. Fully automatic segmentation is still very challenging and difficult to accomplish. Many automatic approaches are domain-dependant, usually applied in the medical field \citep{Christ2016,Moeskops2016,Avendi2016,Bozkurt2018,MartinezMunoz2016,PatinoCorrea2014}. Therefore interactive image segmentation, in which a user supplies some information regarding the objects of interest, is experiencing increasing interest in the last decades  \citep{Boykov2001,Rother2004,Blake2004,Grady2006,Ding2010,Price2010,Li2010,Artan2011,Ding2012,Ducournau2014,Breve2015IJCNN, Breve2015ICCSA,Oh2017,Wang2018b,Liew2017,Wang2018,Lin2016,Dong2016,Wang2016,Wang2016b}.

The user interaction may take place in different ways, depending on the choice of the method, including loosely tracing the desired boundaries \citep{Blake2004,Wang2007}, marking parts of the object(s) of interest and/or background \citep{Boykov2001,Li2004,Grady2006,Price2010,Breve2015IJCNN,Breve2015ICCSA}, loosely placing a bounding box around the objects of interest \citep{Rother2004,Lempitsky2009,Pham2010}, among others. In all scenarios, the goal is to allow the user to select the desired objects with minimal effort \citep{Price2010}.

This paper focuses on the second type of approach, in which the user ``scribbles'' some lines on the object(s) of interest and the background.  The ``scribbles'' are then used as seeds to guide the iterative segmentation process. That is a popular approach because it requires only a quicker and less precise input from the user. They can loosely mark broader interior regions instead of finely tracing near borders \citep{Price2010}.

Graph-cuts is one of the most popular approaches to seeded segmentation, with numerous methods proposed \citep{Boykov2001,Boykov2006,Rother2004,Blake2004,Price2010,Vicente2008}. In graph theory, a cut is a partition of the vertices of a graph into two disjoint subsets \citep{Narkhede2013}. These methods combine explicit edge-finding and region-modeling components, modeled as an optimization problem of minimizing a cut in a weighted graph partitioning foreground and background seeds \citep{Price2010}.

Other approaches rely on graph-based machine learning \citep{Grady2006,Wang2007b,Duchenne2008,Ducournau2014,Breve2015IJCNN,Breve2015ICCSA,Oh2017,Wang2018b,Dong2016,Wang2016,Wang2016b}, where the image is modeled as an affinity graph, where edges encode similarity between neighboring pixels. The segmentation problem may be modeled as an energy function minimization, where the target function is smooth concerning the underlying graph structure \citep{Ducournau2014}. Some deep-learning approaches were also recently proposed \citep{Liew2017,Wang2018,Lin2016}.

The emergence of graph-based techniques is also due to the development of complex networks theory. In the last decades, the network research moved from small graphs to the study of statistical properties of large-scale graphs. It was discovered that a regular network diameter might be drastically reduced by randomly changing a few edges while preserving its local structure, measured by clustering coefficient \citep{Watts1998}. The resulting networks are called \emph{small-world networks} and they represent some real networks, like social and linguistics networks. Small-world networks have tightly interconnected clusters of nodes and a shortest mean path length that is similar to a random graph with the same number of nodes and edges \citep{Humphries2008}.

This paper introduces a new graph-based method for interactive segmentation. It is simpler than many other methods. It does not incorporate any specific edge-finding or region-modeling components. There is also no explicit optimization process. The graphs are merely used to propagate the labels from the ``scribbles'' to unlabeled pixels iteratively, directly through the weighted edges. The method has two consecutive stages. In the first stage, a $k$-nearest neighbor ($k$-NN) graph is built based on the similarity among all pixels on a reduced version of the input image, with each node representing a pixel (a group of pixels of the original image). In the second stage, the full-size image is used, a new graph is built with each node representing a single pixel, which is connected only to the nodes representing the $8$ adjacent pixels in the image. The propagation occurs only to the nodes that were not confidently labeled during the first stage.

The propagation approach has some similarities with that proposed by \citet{Wang2007b}. However, the graph construction is fundamentally different, as in the first stage nodes are not connected in a grid, but rather based on the color components and location of the pixels they represent. In this sense, the label propagation is faster, as the graph usually presents the small-world property of complex networks \citep{Watts1998}.

The graph construction phase share some similarities with that proposed by \citet{Breve2017ICCSA}. However, that approach uses undirected and unweighted graphs while the current study uses weighted digraphs. The propagation approach is also completely different. That model uses particles walking through the graph to propagate label information, in a nature-inspired approach of competition and cooperation for territory. The proposed method approach is much faster, as the label information spreads directly through the graph. Finally, the particles model is stochastic and this proposed model is deterministic.

In spite of its simplicity, the proposed method can perform interactive image segmentation with high accuracy. It was applied to the $50$ images from the Microsoft GrabCut dataset \citep{Rother2004} and the mean error rate achieved is comparable to those obtained by some state-of-the-art methods. Moreover, its computational complexity order is only linear, $O(n)$, where $n$ is the amount of the pixels in the image in the best case scenario, and linearithmic, $O(n \log n)$, in the worst case. It can also be applied to multi-class problems at no extra cost.

The remaining of this paper is organized as follows. Section \ref{sec:ModelDescription} describes the proposed model. Section \ref{sec:ComputerSimulations} presents some computer simulations to show the viability of the method.  Section \ref{sec:CompComplex} discuss the time and storage complexity of the algorithm and the small-world property of its networks. In Section \ref{sec:Benchmark}, the method is applied to the Microsoft GrabCut dataset and its results are compared to those achieved by some state-of-the-art algorithms. Some parameter analysis are also conducted in this section. Finally, the conclusions are drawn on Section~\ref{sec:Conclusions}.

\section{Model Description}
\label{sec:ModelDescription}

The proposed algorithm is divided into two stages. In the first stage, the input image is reduced to one ninth of its original size using bicubic interpolation, and a network is built with each node representing a pixel in the downsized image. The edges among them are built by connecting each node to its $k$-nearest neighbors, in a complex arrangement, which considers both the pixel location and color. Then, label information is propagated iteratively through this network. Usually, most pixels are labeled with confidence in this stage.

In the second stage, the full input image is used. Again, each node represents a single pixel. However, this time, the connections are made only from the pixels not confidently labeled in the first stage to the nodes representing the adjacent pixels in the image, in a grid arrangement, which considers only pixel location. Label information propagates iteratively again, only to the unlabeled nodes. Therefore, the remaining pixels are labeled at this stage.

In both networks, the same set of pixel features, considering both color and location, are extracted to define the edge weights. The whole procedure is detailed in the following subsections.

\subsection{The First Stage}
\label{sec:FirstStage}

In the first stage, the input image is resized to one ninth of its original size (one third in each dimension) using bicubic interpolation. Then, the set of pixels of the resized image are reorganized as $\mathfrak{X} = \{x_1,x_2,\dots,x_L,x_{L+1},\dots,x_N\}$, such that $\mathfrak{X}_{L} = \{x_i\}_{i=1}^{L}$ is the labeled pixel subset and $\mathfrak{X}_{U} = \{x_i\}_{i=L+1}^{N}$ is the unlabeled pixels set. $\mathfrak{L} = \{1,\dots,C\}$ is the set containing the labels. $y: \mathfrak{X} \rightarrow \mathfrak{L}$ is the function associating each $x_i \in \mathbf{\chi}$ to its label $y(x_i)$. The proposed model estimates $y(x_i)$ for each unlabeled pixel $x_i \in \mathfrak{X}_{U}$.

The labels are extracted from an image with the user input (``scribbles''), in which a different color represents each class, and another color is used for the unlabeled pixels. In the first stage, this image is also resized to one ninth of its original size, but using the nearest-neighbor interpolation; otherwise, new colors would be introduced and mistakenly interpreted as new classes.

\subsubsection{Graph Generation}
\label{sec:GraphGeneration}

For each pixel $x_i$, a set of nine features are extracted. They are shown in Table~\ref{tab:Features}.

\begin{table}
\centering
\caption{List of features extracted from each image to be segmented}
\begin{tabular}{rl}
  \toprule
  \# & Feature Description \\
  \midrule
  1 & Pixel row location \\
  2 & Pixel column location \\
  3 & Red (R) component of the pixel \\
  4 & Green (G) component of the pixel \\
  5 & Blue (B) component of the pixel \\
  6 & Value (V) component of the pixel from a RGB to HSV transform\\
  7 & Excess Red Index (ExR) of the pixel \\
  8 & Excess Green Index (ExG) of the pixel \\
  9 & Excess Blue Index (ExB) of the pixel \\
  \bottomrule
\end{tabular}
\label{tab:Features}
\end{table}

The V component ($6$) is obtained from the RGB components using the method described by \citet{Smith1978}. ExR, ExG, and ExB ($7$ to $9$) indexes are obtained from the RGB components as described in the ``Image Segmentation Data Set''\footnote{Available at \url{http://archive.ics.uci.edu/ml/datasets/image+segmentation}} \citep{Dua2017}:
\begin{eqnarray}
  ExR &=& (2R - (G + B))  \\
  ExG &=& (2G - (R + B))  \\
  ExB &=& (2B - (G + R))
\end{eqnarray}
The Excess Green Index (ExG) and some of its derivatives are commonly employed on segmentation of agricultural images \citep{Guijarro2011}. These indexes are useful for identifying the amount of a color component concerning the others. In this paper, they are used because they decrease the distance among pixels representing the same segment which may have different amounts of incident light.

This set of features was chosen based on some earlier experiments with a preliminary version of the algorithm which had a single stage and no image resize step. It was applied to a subset of $9$ images from the GrabCut dataset, with a set of $23$ features, including H and S components obtained with the method described by \citet{Smith1978}, and mean (M) and standard deviation (SD) of RGB and HSV components on a $3 \times 3$ window around the pixel. All features were normalized to have zero mean and unit variance. After that, each feature had a weight to be used in the calculation of the distance among pixels. The weights of the $23$ features were optimized using the Genetic Algorithm from the MATLAB Global Optimization Toolbox, with its default parameters and a fitness function to minimize the error rate, given by the number of mislabeled pixels in relation to all unlabeled pixels. Based on the results shown in Table~\ref{tab:FeatureStudy}, H and S features were discarded because of their low relevance in most images. The Mean and Standard deviation based features were discarded because the current version of the algorithm works on the resized image, so each pixel is already roughly an average of a $3 \times 3$ window around the pixel. The remaining features are those presented in Table~\ref{tab:Features}.

\begin{table}
\caption{Preliminary study on a larger feature set with $9$ images from the GrabCut dataset and weights optimized using a Genetic Algorithm.}
\resizebox{\textwidth}{!}{%
\begin{tabular}{ccccccccccccc}
\hline
{\bf Image /
Feature} &  {\bf dog} & {\bf 21077} & {\bf 124084} & {\bf 271008} & {\bf 208001} & {\bf llama} & {\bf doll} & {\bf person7} & {\bf sheep} & {\bf teddy} & \multicolumn{ 2}{c}{{\bf Mean}} \\
\hline
 {\bf Row} &     $0.99$ &     $0.96$ &     $0.99$ &     $0.44$ &     $0.98$ &     $0.23$ &     $0.97$ &     $0.79$ &     $0.94$ &     $0.44$ & {\bf $0.77$} & {\bf $(\pm 0.29)$} \\

 {\bf Col} &     $0.80$ &     $0.72$ &     $0.67$ &     $0.91$ &     $0.97$ &     $1.00$ &     $0.34$ &     $0.99$ &     $0.83$ &     $0.39$ & {\bf $0.76$} & {\bf $(\pm 0.24)$} \\

   {\bf R} &     $0.63$ &     $0.03$ &     $0.06$ &     $0.91$ &     $0.45$ &     $0.45$ &     $0.56$ &     $0.58$ &     $0.88$ &     $0.28$ & {\bf $0.48$} & {\bf $(\pm 0.30)$} \\

   {\bf G} &     $0.35$ &     $0.49$ &     $0.25$ &     $0.96$ &     $0.30$ &     $0.28$ &     $0.31$ &     $0.10$ &     $0.54$ &     $0.67$ & {\bf $0.42$} & {\bf $(\pm 0.25)$} \\

   {\bf B} &     $0.96$ &     $0.45$ &     $0.18$ &     $0.36$ &     $0.54$ &     $0.46$ &     $0.26$ &     $0.19$ &     $0.80$ &     $0.39$ & {\bf $0.46$} & {\bf $(\pm 0.25)$} \\

   {\bf H} &     $0.03$ &     $0.04$ &     $0.08$ &     $0.37$ &     $0.07$ &     $0.74$ &     $0.03$ &     $0.15$ &     $0.11$ &     $0.14$ & {\bf $0.17$} & {\bf $(\pm 0.22)$} \\

   {\bf S} &     $0.74$ &     $0.10$ &     $0.06$ &     $0.09$ &     $0.10$ &     $0.30$ &     $0.21$ &     $0.07$ &     $0.10$ &     $0.92$ & {\bf $0.27$} & {\bf $(\pm 0.31)$} \\

   {\bf V} &     $0.65$ &     $0.28$ &     $0.71$ &     $0.94$ &     $0.80$ &     $0.24$ &     $0.18$ &     $0.61$ &     $0.74$ &     $0.29$ & {\bf $0.54$} & {\bf $(\pm 0.27)$} \\

 {\bf ExR} &     $0.96$ &     $0.21$ &     $0.06$ &     $0.85$ &     $0.84$ &     $0.79$ &     $0.13$ &     $0.38$ &     $0.12$ &     $0.47$ & {\bf $0.48$} & {\bf $(\pm 0.35)$} \\

 {\bf ExB} &     $0.72$ &     $0.36$ &     $0.64$ &     $0.70$ &     $0.42$ &     $0.47$ &     $0.16$ &     $0.83$ &     $0.35$ &     $0.98$ & {\bf $0.56$} & {\bf $(\pm 0.25)$} \\

 {\bf ExG} &     $0.75$ &     $0.10$ &     $0.07$ &     $0.22$ &     $0.19$ &     $0.04$ &     $0.78$ &     $0.29$ &     $0.24$ &     $0.32$ & {\bf $0.30$} & {\bf $(\pm 0.26)$} \\

  {\bf MR} &     $0.11$ &     $0.38$ &     $0.03$ &     $0.35$ &     $0.13$ &     $0.21$ &     $0.78$ &     $0.34$ &     $0.81$ &     $0.22$ & {\bf $0.34$} & {\bf $(\pm 0.27)$} \\

  {\bf MG} &     $0.13$ &     $0.23$ &     $0.79$ &     $0.30$ &     $0.76$ &     $0.47$ &     $0.70$ &     $0.18$ &     $0.61$ &     $0.93$ & {\bf $0.51$} & {\bf $(\pm 0.29)$} \\

  {\bf MB} &     $0.49$ &     $0.31$ &     $0.33$ &     $0.42$ &     $0.45$ &     $0.24$ &     $0.15$ &     $0.29$ &     $0.77$ &     $0.80$ & {\bf $0.42$} & {\bf $(\pm 0.21)$} \\

 {\bf SDR} &     $0.01$ &     $0.12$ &     $0.06$ &     $0.11$ &     $0.08$ &     $0.38$ &     $0.02$ &     $0.27$ &     $0.20$ &     $0.22$ & {\bf $0.15$} & {\bf $(\pm 0.12)$} \\

 {\bf SDG} &     $0.01$ &     $0.08$ &     $0.09$ &     $0.03$ &     $0.09$ &     $0.38$ &     $0.21$ &     $0.02$ &     $0.27$ &     $0.06$ & {\bf $0.12$} & {\bf $(\pm 0.12)$} \\

 {\bf SDB} &     $0.00$ &     $0.05$ &     $0.05$ &     $0.04$ &     $0.22$ &     $0.22$ &     $0.06$ &     $0.40$ &     $0.13$ &     $0.01$ & {\bf $0.12$} & {\bf $(\pm 0.13)$} \\

  {\bf MH} &     $0.58$ &     $0.04$ &     $0.16$ &     $0.91$ &     $0.15$ &     $0.92$ &     $0.03$ &     $0.27$ &     $0.14$ &     $0.86$ & {\bf $0.40$} & {\bf $(\pm 0.37)$} \\

  {\bf MS} &     $0.65$ &     $0.31$ &     $0.04$ &     $0.06$ &     $0.21$ &     $0.21$ &     $0.36$ &     $0.89$ &     $0.41$ &     $0.67$ & {\bf $0.38$} & {\bf $(\pm 0.28)$} \\

  {\bf MV} &     $0.02$ &     $0.95$ &     $0.04$ &     $0.35$ &     $0.55$ &     $0.78$ &     $0.66$ &     $0.57$ &     $0.80$ &     $0.25$ & {\bf $0.50$} & {\bf $(\pm 0.32)$} \\

 {\bf SDH} &     $0.17$ &     $0.41$ &     $0.19$ &     $0.39$ &     $0.48$ &     $0.32$ &     $0.07$ &     $0.31$ &     $0.38$ &     $0.08$ & {\bf $0.28$} & {\bf $(\pm 0.14)$} \\

 {\bf SDS} &     $0.03$ &     $0.41$ &     $0.22$ &     $0.15$ &     $0.18$ &     $0.23$ &     $0.13$ &     $0.50$ &     $0.51$ &     $0.02$ & {\bf $0.24$} & {\bf $(\pm 0.18)$} \\

 {\bf SDV} &     $0.61$ &     $0.21$ &     $0.07$ &     $0.10$ &     $0.13$ &     $0.31$ &     $0.03$ &     $0.10$ &     $0.38$ &     $0.24$ & {\bf $0.22$} & {\bf $(\pm 0.18)$} \\
\hline
\end{tabular}
}
\label{tab:FeatureStudy}
\end{table}

In the proposed method, the $9$ features from Table~\ref{tab:Features} are normalized to have zero mean and unit variance. After that, the components may be scaled by a vector of weights $\mathbf{\lambda}$ to emphasize/de-emphasize each feature during the graph generation. However, for simplicity, in all experiments on this paper, only two set of weights were used as $\mathbf{\lambda}$. They will be later referenced as:
\begin{equation}
\begin{array}{ccccccccc}
    \mathbf{\lambda_1} = [1.0 & 1.0 & 1.0 & 1.0 & 1.0 & 1.0 & 1.0 & 1.0 & 1.0] \\
    \mathbf{\lambda_2} = [1.0 & 1.0 & 0.5 & 0.5 & 0.5 & 0.5 & 0.5 & 0.5 & 0.5]
\end{array}
\end{equation}
Thus, $\mathbf{\lambda_1}$ means all features have the same weight, and $\mathbf{\lambda_2}$ means the two location features have more weight than the seven color features. While there are many other possible weight combinations, there is not a reliable method to set them a priori, without relying on the segmentation results.

A directed and weighted graph is created representing the image. It is defined as $\mathbf{G} = (\mathbf{V},\mathbf{E})$, where $\mathbf{V} = \{v_1,v_2,\dots,v_n\}$ is the set of $n$ nodes, and $\mathbf{E}$ is the set of $m$ edges $(v_i, v_j)$. Each node $v_i$ corresponds to a pixel $x_i$. There is an edge between $v_i$ and $v_j$ only if $v_i$ is unlabeled ($x_i \in \mathfrak{X}_{U}$) and $v_j$ is among the $k$-nearest neighbors of $v_i$, considering the Euclidean distance between $x_i$ and $x_j$ features. Along with development, it was noticed that $k=10$ provides reasonable results in most images, as long as representative seeds are provided. But this parameter may be fine-tuned for each specific image to achieve better segmentation results.

For each edge $(v_i, v_j) \in \mathbf{E}$, there is a corresponding weight $W_{i,j}$, which is defined using a Gaussian kernel:
\begin{equation}
\label{eq:Weights}
  W_{i,j} = \exp\frac{-d(x_i,x_j)^2}{2\sigma^2}
\end{equation}
where $d(x_i,x_j)$ is the Euclidean distance between $x_i$ and $x_j$. Along development, it was noticed that $\sigma$ is not a very sensitive parameter. Therefore, $\sigma=0.5$ is fixed for all computer simulations in this paper.

\subsubsection{Label Propagation}
\label{sec:LabelPropagation}

For each node $v_i$, a domination vector $\mathbf{v_i^\omega(t)} = \{v_i^{\omega_1}(t), v_i^{\omega_2}(t), \dots, v_i^{\omega_C}(t) \}$ is created. Each element $v_i^{\omega_c}(t) \in [0, 1]$ corresponds to the domination level from the class $c$ over the node $v_i$. The sum of the domination vector in each node is always constant:
\begin{equation}
    \sum_{c=1}^{C} v_i^{\omega_c} = 1
\end{equation}

Nodes corresponding to labeled pixels are fully dominated by their corresponding class, and their domination vectors never change. On the other hand, nodes corresponding to unlabeled pixels have variable domination vectors. They are initially set in balance among all classes. Thus, for each node $v_i$, each element $v_i^{\omega_c}$ of the domination vector $\mathbf{v_i^\omega}$ is set as follows:
\begin{equation}\label{eq:NodesInit}
    v_i^{\omega_c}(0) = \left\{
    \begin{array}{ccl}
        1 & & \mbox{if $x_i$ is labeled and $y(x_i) = c$} \\
        0 & & \mbox{if $x_i$ is labeled and $y(x_i) \neq c$} \\
        \frac{1}{C} & & \mbox{if $x_i$ is unlabeled}
    \end{array}\right.
\end{equation}

Then, the iterative label propagation process takes place. At each iteration $t$, each unlabeled node gets contributions from all its neighbors to calculate its new domination levels. Therefore, for each unlabeled node $v_i$, the domination levels are updated as follows:
\begin{equation}
\label{eq:NodesUpdate}
    \mathbf{v_i^\omega(t+1)} = \frac{\sum_{j \in N(v_i)} W_{i,j} \mathbf{v_j^\omega(t)}}{\sum_{j \in N(v_i)} W_{i,j}}
\end{equation}
where $N(v_i)$ is the set of $v_i$ neighbors. In this sense, the new domination vector $\mathbf{v_i^\omega}$ is the weighted arithmetic mean of all its neighbors domination vectors, no matter if they are labeled or unlabeled.

The iterative process stops when the domination vectors converge. At this point, $\mathbf{v_i^\omega}$ is re-organized to form a bi-dimensional grid, with each vector element in the same position of its corresponding pixel in the resized image. Then, the grid is enlarged to the size of the original input image, using bilinear interpolation, so $\mathbf{v_i^\omega}$ has a vector for each pixel of the original input image.

When the first stage finishes, most pixels are completely dominated by a single class. The exceptions are usually the pixels in classes borders. Thus, for every node $v_i$, if there is a highly dominant class, that class is assigned to the corresponding pixel:
\begin{equation}
\label{eq:LabelUnlabeledStage1}
    \forall v_i^{\omega_c} = 1, y(x_i) = c
\end{equation}
where $y(x_i)$ is the class assigned to $x_i$. Otherwise, the pixel is left to be labeled in the second stage.

\subsection{The Second Stage}
\label{sec:SecondStage}

In the second stage, nodes that were not labeled in the first stage continue to receive contributions from their neighbors. However, in the second stage a new graph is built, in which every pixel in the input image becomes a node (no resizing), and each node $v_i$ corresponding to an unlabeled pixel ($x_i \in \mathfrak{X}_{U}$) is connected to the $8$ nodes $v_j$ representing the adjacent pixels in the original image, except for pixels in the image borders, which have only $3$ or $5$ adjacent pixels. So, in the second phase, neighbors are defined only by location, but the edge weights are still defined by Eq. \eqref{eq:Weights}, using all the nine features.

Notice that the domination vectors $\mathbf{v_i^\omega}$ are not reset before the second stage. The iterative label propagation process in the second stage also uses Eq. \eqref{eq:NodesUpdate}, and it stops when the domination vectors converge. At this point, all the still unlabeled pixels are labeled after the class that dominated their corresponding node:
\begin{equation}
\label{eq:LabelUnlabeledStage2}
    y(x_i) = \arg\max_c v_i^{\omega_c}
\end{equation}
where $y(x_i)$ is the class assigned to $x_i$.

\subsection{Stop Criterion}

In both stages, the convergence is measured through the average maximum domination level, which is defined as follows:
\begin{equation}\label{eq:StopCrit}
    \langle v_i^{\omega_{m}} \rangle, m=\arg\max_c v_i^{\omega_c}
\end{equation}
for all $v_i$ representing unlabeled nodes. $\langle v_i^{\omega_{m}} \rangle$ is checked every $10$ iterative steps and the iterations stop when the increase is below $\omega$ between two consecutive checkpoints. In this paper, $\omega=10^{-4}$ is used in all computer simulations.

\subsection{The Algorithm}
\label{sec:Algorithm}

Overall, the proposed algorithm can be outlined as shown in Algorithm \ref{alg:Algorithm}.

\begin{algorithm}[h] \small
    Downsize the input image with bicubic interpolation, as described in Subsection \ref{sec:FirstStage}\;
    Build a directed $k$-NN digraph for the downsized image, as described in Subsection \ref{sec:GraphGeneration}\;
    Define the edge weights using Eq. \eqref{eq:Weights}\;
    Set nodes' domination levels by using Eq. \eqref{eq:NodesInit}\;
    \Repeat{convergence of the domination levels}
    {
        \For{each unlabeled node}
        {
            Update node domination levels by using Eq. \eqref{eq:NodesUpdate}\;
        }
    }
    Label unlabeled pixel where there is a highly dominant class using Eq. \eqref{eq:LabelUnlabeledStage1}\;
    Enlarge the domination levels matrix with bilinear interpolation, as described in Subsection \ref{sec:LabelPropagation}\;
    Build a directed graph for the original input image, using a grid arrangement, as described in Subsections \ref{sec:SecondStage} and \ref{sec:GraphGeneration}\;
    Define the edge weights using Eq. \eqref{eq:Weights}\;
    \Repeat{convergence of the domination levels}
    {
        \For{each unlabeled node}
        {
            Update node domination levels by using Eq. \eqref{eq:NodesUpdate}\;
        }
    }
    Label remaining unlabeled pixels using Eq. \eqref{eq:LabelUnlabeledStage2}\;
    \caption{The proposed method algorithm}
    \label{alg:Algorithm}
\end{algorithm}

\section{Computer Simulations}
\label{sec:ComputerSimulations}

\begin{figure}
\centering
\setlength\tabcolsep{1.5pt}
\begin{tabular}{ccc}
\subfloat{\includegraphics[height=2.8cm]{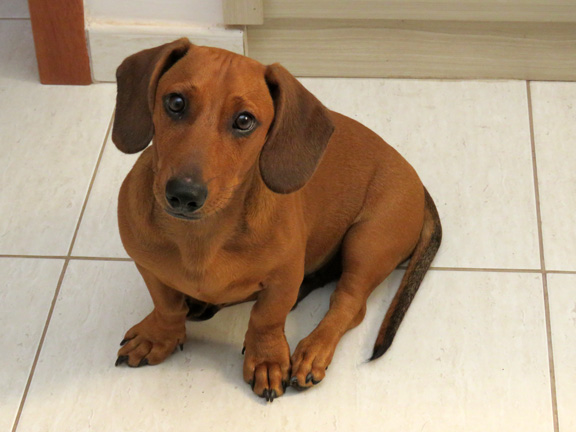}                    } &
\subfloat{\includegraphics[height=2.8cm]{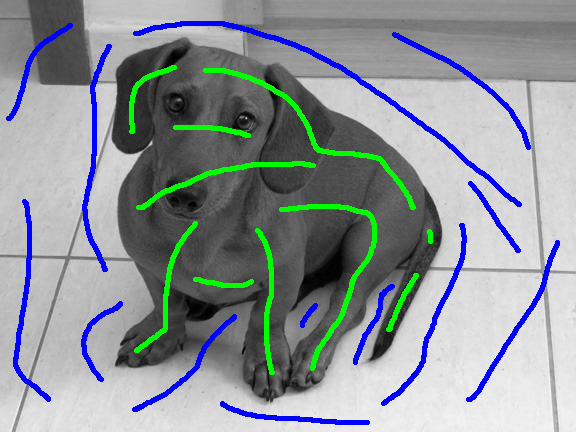}        } &
\subfloat{\includegraphics[height=2.8cm]{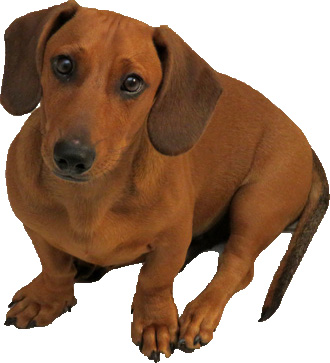}             } \\
\subfloat{\includegraphics[height=2.8cm]{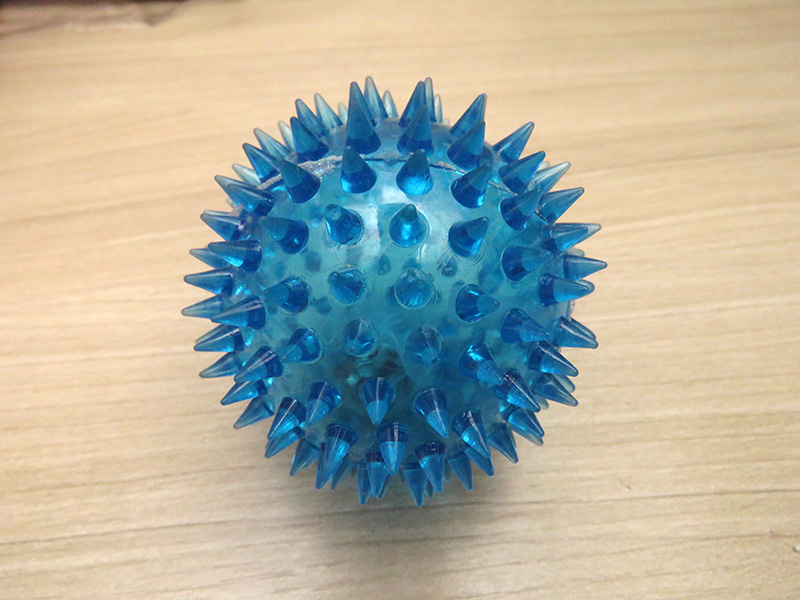}                    } &
\subfloat{\includegraphics[height=2.8cm]{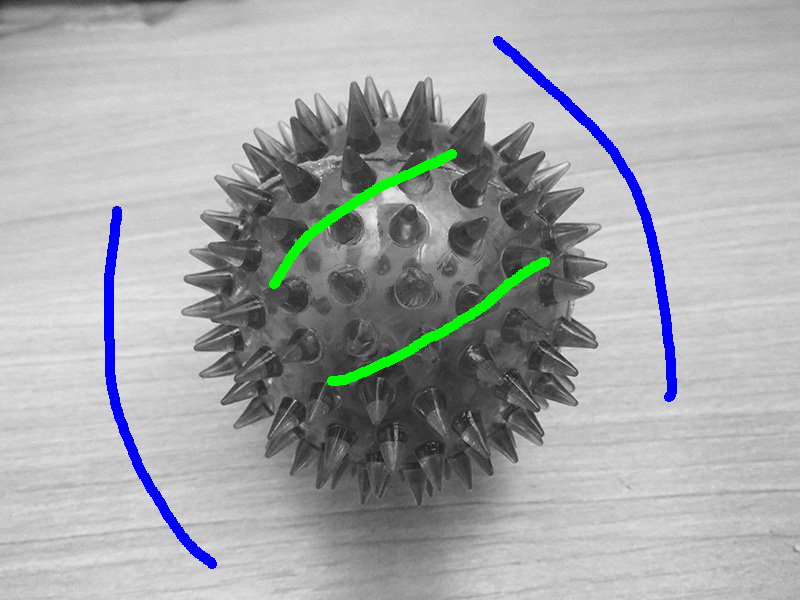}        } &
\subfloat{\includegraphics[height=2.8cm]{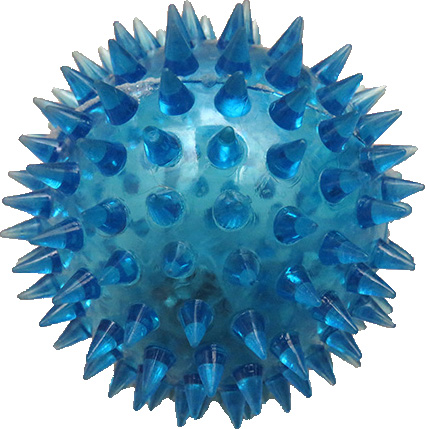}             } \\
\subfloat{\includegraphics[height=2.8cm]{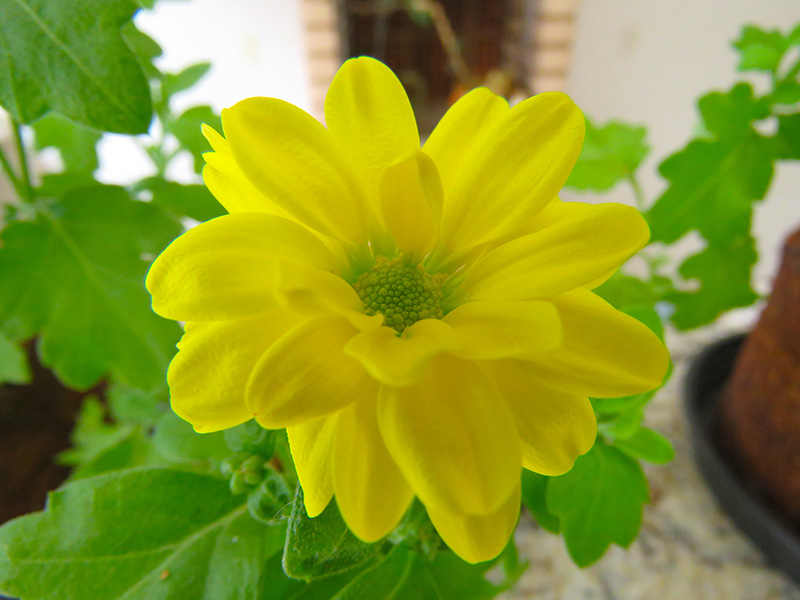}                    } &
\subfloat{\includegraphics[height=2.8cm]{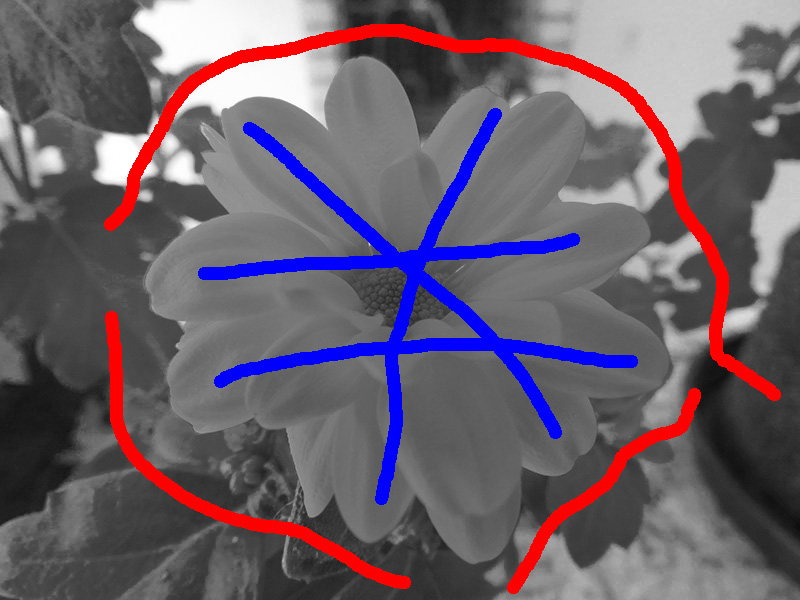}        } &
\subfloat{\includegraphics[height=2.8cm]{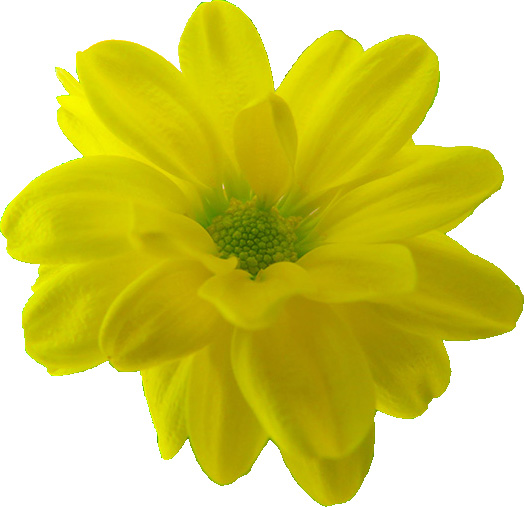}             } \\
\subfloat{\includegraphics[height=2.8cm]{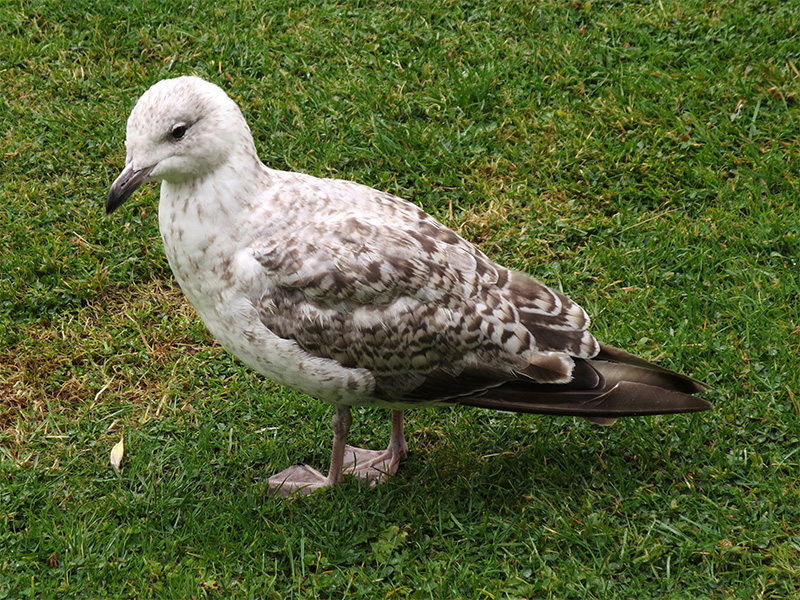}                    } &
\subfloat{\includegraphics[height=2.8cm]{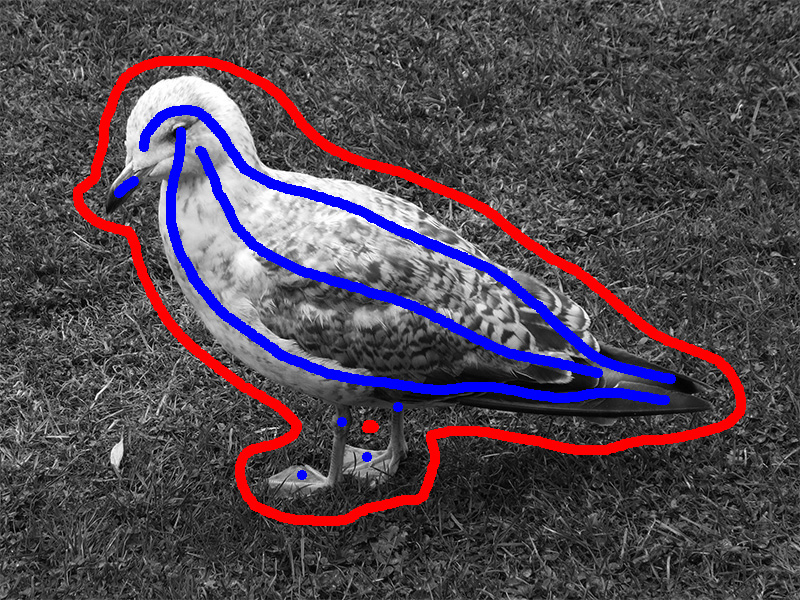}        } &
\subfloat{\includegraphics[height=2.8cm]{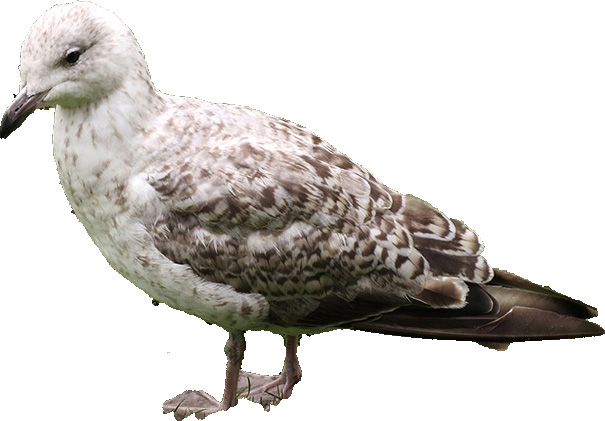}             } \\
\subfloat{\includegraphics[height=2.8cm]{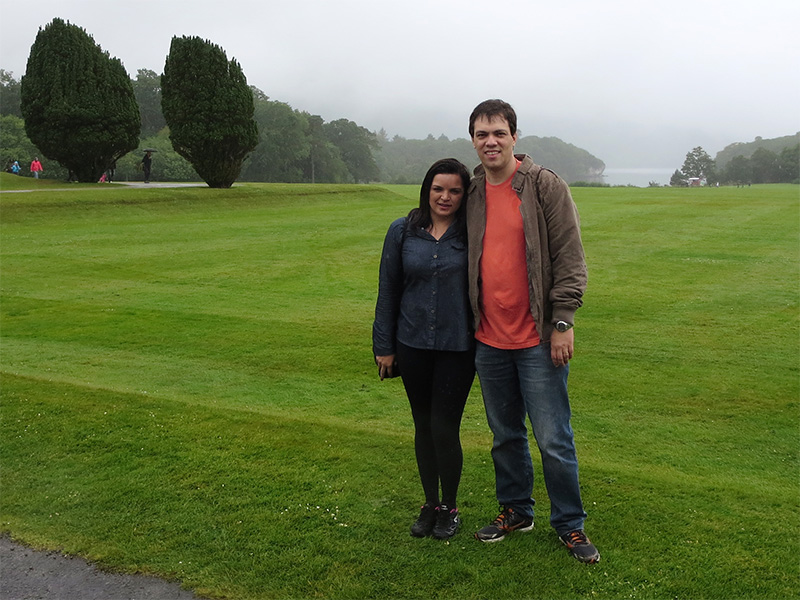}                    } &
\subfloat{\includegraphics[height=2.8cm]{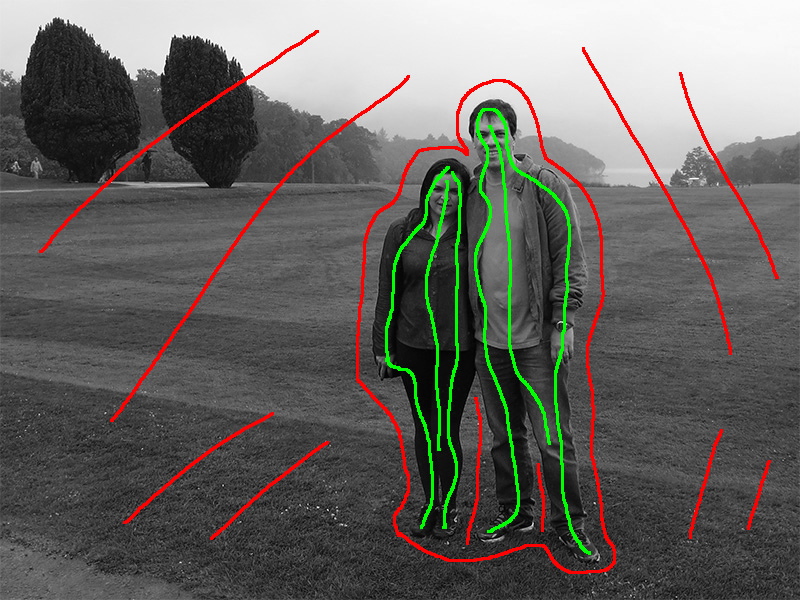}        } &
\subfloat{\includegraphics[height=2.8cm]{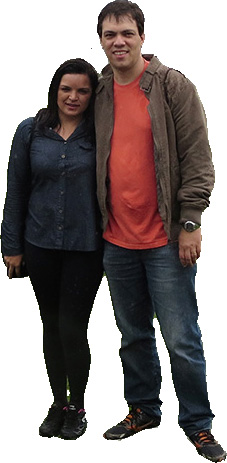}             } \\
(a) & (b) & (c)
\end{tabular}
\caption{Foreground and background segmentation by the proposed method: (a) the real-world images to be segmented, (b) the ``scribbles'' provided by the user, and (c) the segmentation results.}
\label{fig:TwoClass}
\end{figure}

\begin{figure}
\centering
\setlength\tabcolsep{2pt}
\begin{tabular}{cc}
\subfloat{\includegraphics[height=4.2cm]{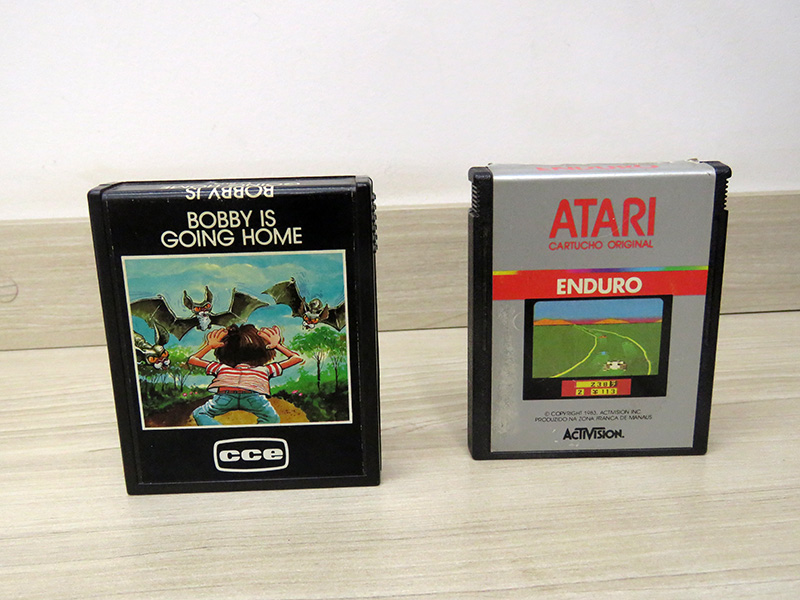}                    } &
\subfloat{\includegraphics[height=4.2cm]{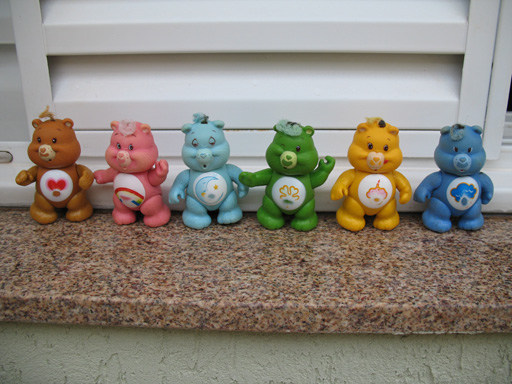}                    } \\
\multicolumn{2}{c}{(a)} \\
\subfloat{\includegraphics[height=4.2cm]{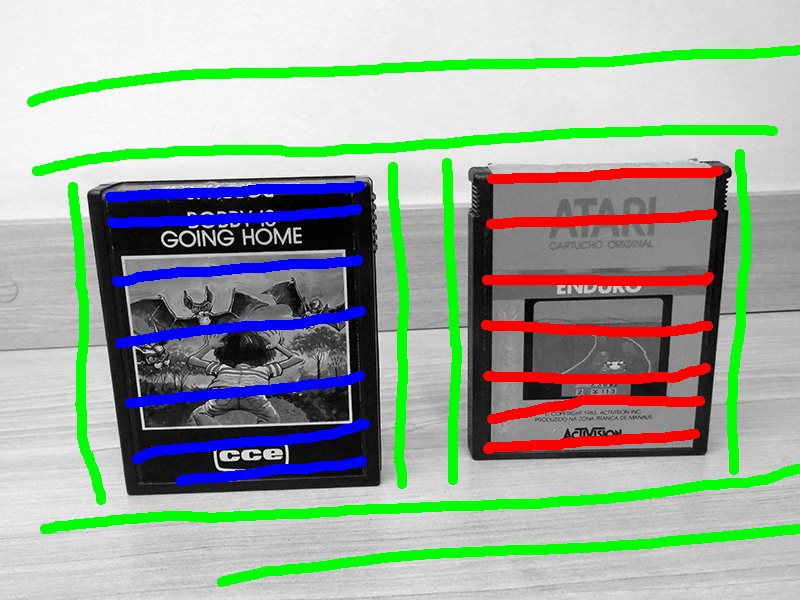}        } &
\subfloat{\includegraphics[height=4.2cm]{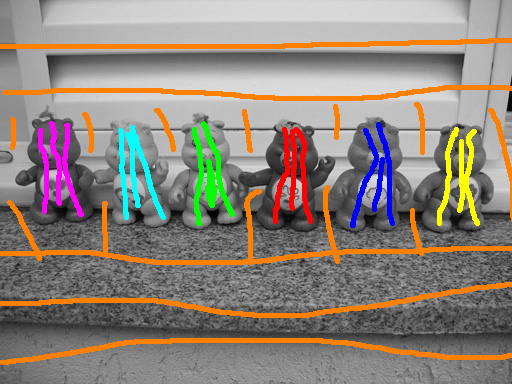}        } \\
\multicolumn{2}{c}{(b)} \\
    \begin{tabular}{cc}
    \subfloat{\includegraphics[height=3cm]{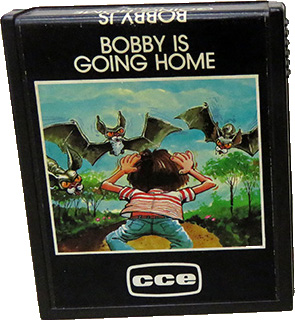}} &
    \subfloat{\includegraphics[height=3cm]{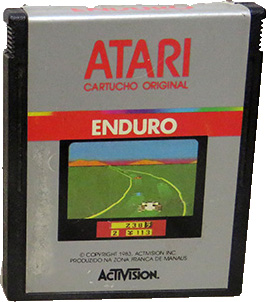}} \\
    \end{tabular}
&
    \begin{tabular}{ccc}
    \subfloat{\includegraphics[height=2cm]{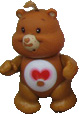}} &
    \subfloat{\includegraphics[height=2cm]{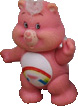}} &
    \subfloat{\includegraphics[height=2cm]{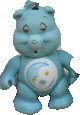}} \\
    \subfloat{\includegraphics[height=2cm]{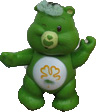}} &
    \subfloat{\includegraphics[height=2cm]{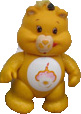}} &
    \subfloat{\includegraphics[height=2cm]{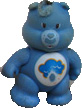}} \\
    \end{tabular}
\\
\multicolumn{2}{c}{(c)}
\end{tabular}
\caption{Multi-class segmentation by the proposed method: (a) the real-world images to be segmented, (b) the ``scribbles'' provided by the user, and (c) the segmentation results.}
\label{fig:MultiClass}
\end{figure}

In this section, some experimental results using the proposed model are presented to show its efficacy in the interactive image segmentation task. First, five real-world images were selected to show that the algorithm can split foreground and background. Later, other two real-world images were selected to show the algorithm results segmenting multiple objects at once. For all images, the parameters are set to their default values, except for $k$ and $\mathbf{\lambda}$. $k$ is tested with some values in the interval $[1 \quad 250]$ and $\mathbf{\lambda}$ is tested with $\mathbf{\lambda_1}$ and $\mathbf{\lambda_2}$. Then, the values that produced the best segmentation results are used for each image. Figure \ref{fig:TwoClass} shows: (a) the five images selected to show the segmentation in background and foreground; (b) the ``scribbles'' that represent the user input, which is shown in different colors for background and foreground, over the gray-scale image; and (c) the segmentation results achieved using the proposed method, shown as the foreground extracted from the background. Figure \ref{fig:MultiClass} shows: (a) the two images selected to show the multi-class segmentation capabilities of the proposed method; (b) the ``scribbles'' representing the user input, which are shown in different colors for each object and the background; and (c) the segmentation results achieved using the proposed method, with each object shown separately. Table \ref{tab:ImageInfo} shows the image sizes and the parameters $\mathbf{\lambda}$ and $k$ used for each of them.

\begin{table}
  \centering
  \caption{Image sizes and parameters used in the segmentation task.}
    \begin{tabular}{lccc}
    \toprule
    \textbf{Image} & \textbf{Size} & \textbf{$\lambda$} & \textbf{$k$} \\
    \midrule
    Dog           & $576 \times 432$ & $\mathbf{\lambda_1}$ & $120$ \\
    Ball          & $800 \times 600$ & $\mathbf{\lambda_1}$ & $15$ \\
    Flower        & $800 \times 600$ & $\mathbf{\lambda_1}$ & $109$ \\
    Bird          & $800 \times 600$ & $\mathbf{\lambda_2}$ & $8$ \\
    Couple        & $800 \times 600$ & $\mathbf{\lambda_2}$ & $14$ \\
    Cartridges    & $800 \times 600$ & $\mathbf{\lambda_2}$ & $11$ \\
    Care Bears    & $512 \times 384$ & $\mathbf{\lambda_2}$ & $15$ \\
    \bottomrule
    \end{tabular}%
  \label{tab:ImageInfo}%
\end{table}%

Notice that the algorithm receives the ``scribbles'' in a different image or layer than the image to be segmented. It considers that each color represents a different segment to be discovered, so the user seeds must be in different colors for each segment. The images to be segmented were added in black-and-white as background to the ``scribbles'' in Figures \ref{fig:TwoClass} and \ref{fig:MultiClass} for illustrative purposes only.

By visually analyzing the segmentation results, one can notice that the proposed method was able to interactively segment different kinds of real-world images, with few mistakes.

\section{Computational Time and Storage Complexity}
\label{sec:CompComplex}

In this section, time and storage complexity order analysis of the algorithm presented in Subsection \ref{sec:Algorithm} are provided.

\subsection{Computational Time Complexity}
\label{sec:CompTimeComplexity}

At the beginning of the Algorithm \ref{alg:Algorithm}, step 1 consists in resizing the input image with bicubic interpolation. This step has complexity order $O(n)$, where $n$ is the number of pixels in the image. Step 2 consists in building a $k$-NN graph. It is possible to find nearest neighbors in logarithmic time using \emph{k}-d trees \citep{Friedman1977}. Therefore, this step computational complexity is $O(n \log n)$. Step 3 calculates edges weights. This step depends on the number of edges. Each node has $k$ edges; therefore, the computational complexity is $O(nk)$. Step 4 is the initialization of the nodes domination levels, and it depends on the number of nodes and classes. Therefore, its computational complexity is $O(nc)$. These first steps are dominated by step 2, which is the network construction as $k$ and $c$ are usually much smaller than $n$.

Then we have the loops from steps 5 to 8. The instruction on step 8 consists in updating the domination levels on a node. Since each node takes contributions from its neighbors, the computational complexity order is $O(k)$. The inner loop executes step 7 for all unlabeled nodes. Most nodes are unlabeled; therefore the inner loop complexity order is $O(nk)$. The outer loop is executed until the algorithm converges. The convergence depends on the network size and connectivity, which are related to $n$ and $k$. Therefore, a set of experiments is performed, first with increasing image sizes and fixed $k$, and later with fixed image size and increasing $k$, to discover how they impact the number of outer loop executions. These will be presented and discussed later.

Step 9 consists of checking domination levels and labeling some nodes. Step 10 increases the domination levels matrix using bilinear interpolation. Both these steps have complexity order $O(nc)$, due to the domination levels matrix size. In step 11, another graph is built, but only adjacent nodes are connected, and to exactly $8$ other nodes (except for nodes representing image border pixels), so this step has complexity order $O(n)$. Step 12 is similar to Step 3, but this time the average node degree is nearly constant, so the complexity order is $O(n)$.

From step 13 to 16, there is another pair of loops. Step 15 runs in constant time $O(1)$, since all nodes (except those representing image border pixels) have the same degree ($8$) no matter the image size. The inner loops execute step 15 for each unlabeled node.  In most cases, there is only a small amount of unlabeled nodes at this point. The outer loop also depends on how many nodes are still unlabeled in the second stage and also on the network connectivity. In the typical scenario, there are few unlabeled nodes, and they form isolated subgraphs. Though it is difficult to calculate the exact computational complexity of the second stage, it is lower than $O(n \log n)$ in any typical case. The set of experiments also measures the number of outer loop iterations in the second stage, as $n$ and $k$ increases.

Finally, step 17 is similar to step 9. It also has complexity order $O(nc)$.

\begin{figure}
\centering
\setlength\tabcolsep{1pt}
\begin{tabular}{cc}
    \subfloat{\includegraphics[height=3.8cm]{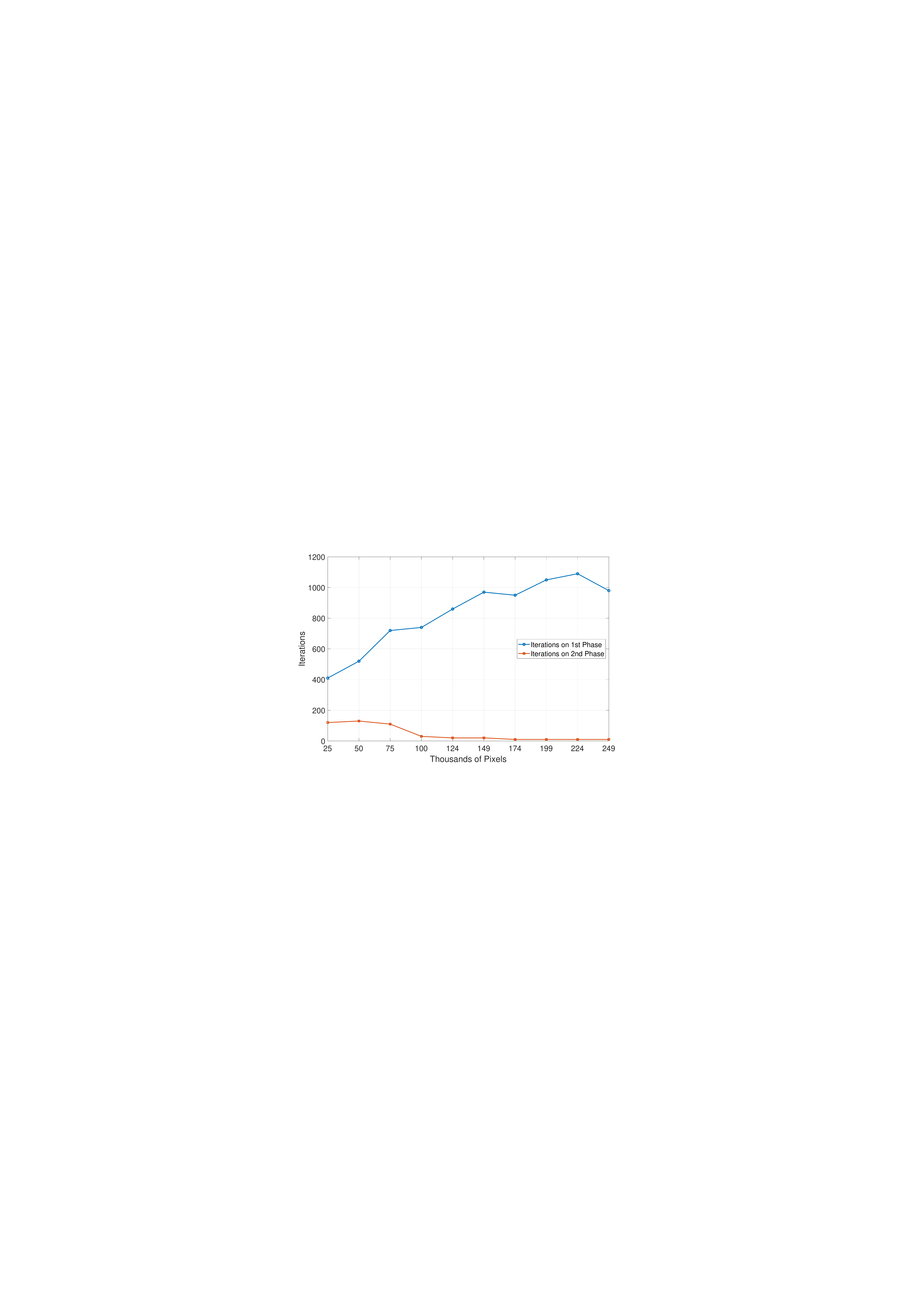}                    } &
    \subfloat{\includegraphics[height=3.8cm]{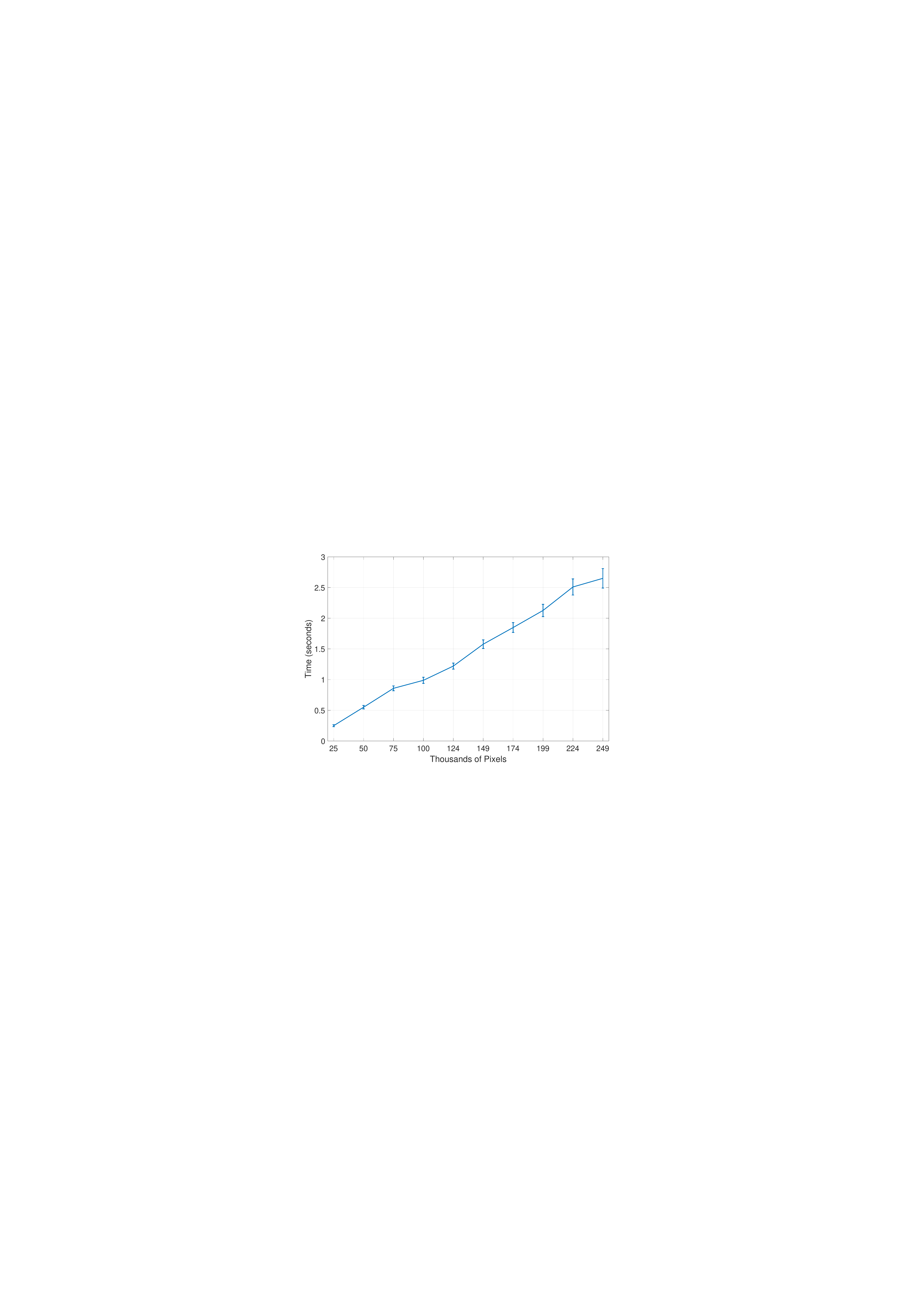}                    } \\
    \multicolumn{2}{c}{(a)} \\
    \subfloat{\includegraphics[height=3.8cm]{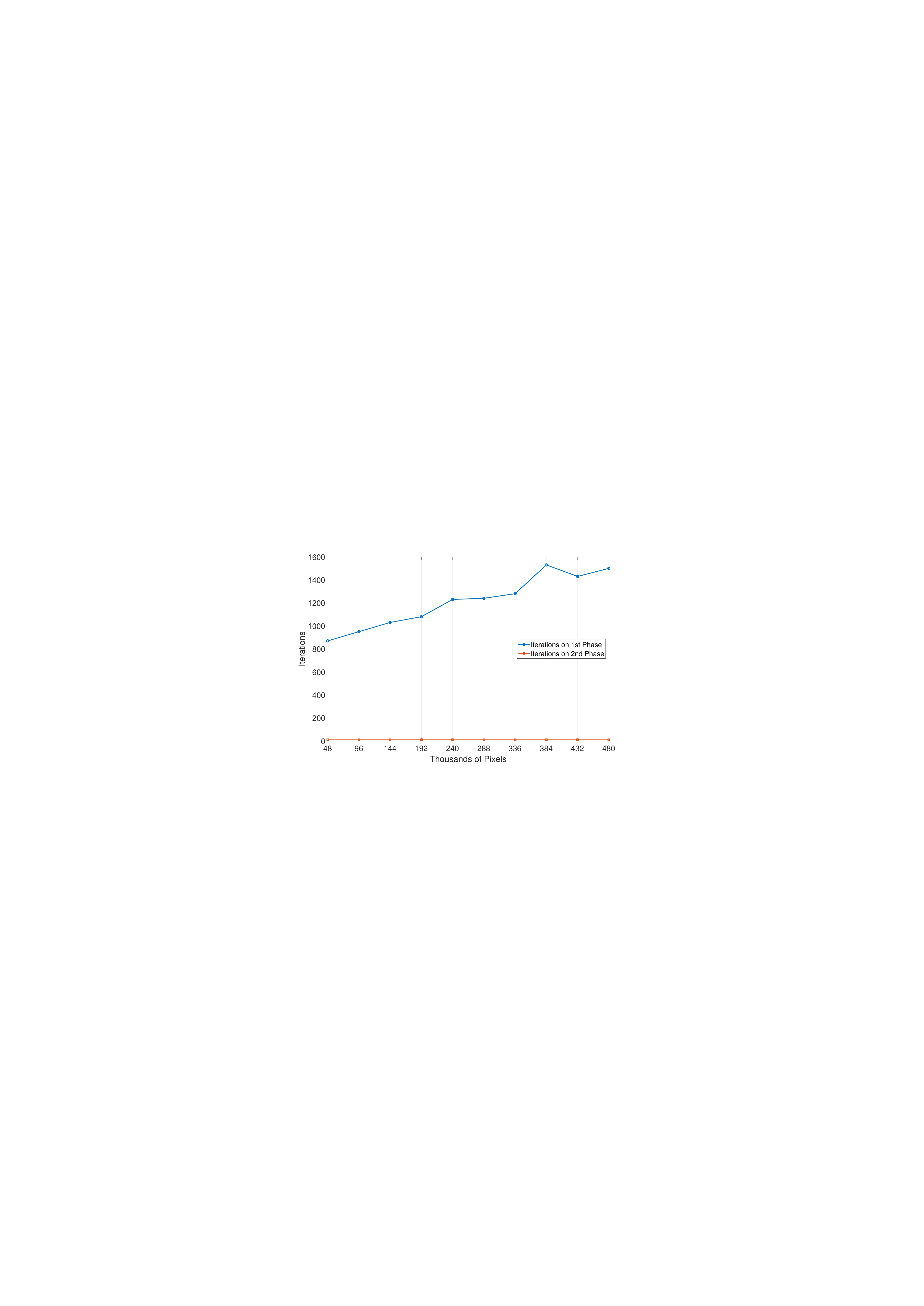}                    } &
    \subfloat{\includegraphics[height=3.8cm]{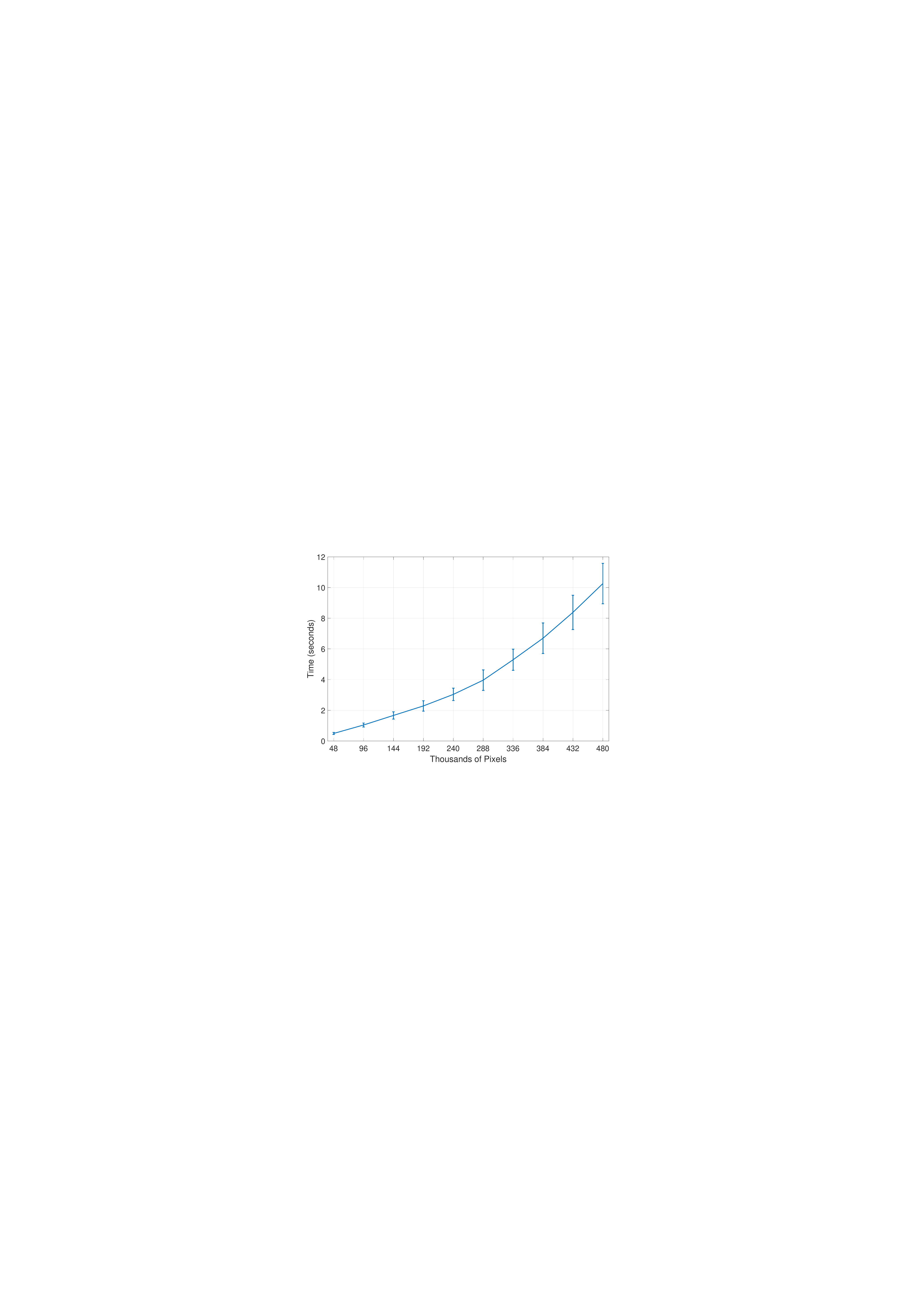}                    } \\
    \multicolumn{2}{c}{(b)} \\
\end{tabular}
\caption{Number of iterations (left) and time (right) required to the convergence, with images with $10\%$ to $100\%$ of their original size and $k=10$. (a) ``Dog'' image. (b) ``Bird'' image. Each point in time traces is the average of $100$ realizations. The error bars represent standard deviation.}
\label{fig:complexkfix}
\end{figure}

Figure \ref{fig:complexkfix} and Tables \ref{tab:ComplexRalphKfix} and \ref{tab:ComplexBirdKfix} show the number of iterations of the outer loops of the first and second stages, and time required to convergence when the proposed method is applied to two images from Figure \ref{fig:TwoClass}: ``Dog'' and ``Bird'' (which are in the first and the fourth row, respectively). Each image and their respective ``scribbles'' images are resized to $10\%, 20\%, \ldots, 100\%$ of their original size, while $k=10$ is kept fixed. By analyzing the graphics, it is possible to realize that as the image size increases, the execution time increase is close to linear ($O(n)$). While the first stage inner loop increases linearly on $n$, the outer loop does not increase significantly, which is expected due to the small-world property of complex networks that keep the nodes grouped in clusters, so the label spreading rate does not change much. The second stage requires only a few iterations of the outer loop, only $10$ in most cases, which is the minimum value since the convergence check is performed every $10$ iteration. Tables \ref{tab:ComplexRalphKfix} and \ref{tab:ComplexBirdKfix} also show the error rate in each scenario, which is the number of mislabeled pixels in relation to all the pixels labeled by the algorithm. Notice that the algorithm labels all the image pixels, except those already covered by the ``scribbles''.

\begin{table}
\caption{Amount of iterations on each algorithm phase, execution times, and error rates on the ``Dog'' image with $10\%$ to $100\%$ of its original size and $k=10$. Each configuration is executed $100$ times to get the mean and standard deviation of the execution times.}
\resizebox{\textwidth}{!}{%
\begin{tabular}{ccccccccc}
\hline
{\bf Image Size} & {\bf Width} & {\bf Height} & {\bf Tot. Pixels} & {\bf Ph. 1} & {\bf Ph. 2} & \multicolumn{ 2}{c}{{\bf Time (s)}} & {\bf Error Rate} \\
\hline
    $10\%$ &      $183$ &      $137$ &   $25,071$ &      $410$ &      $120$ &     $0.25$ & $(\pm 0.02)$ &   $0.0018$ \\

    $20\%$ &      $258$ &      $194$ &   $50,052$ &      $520$ &      $130$ &     $0.55$ & $(\pm 0.03)$ &   $0.0021$ \\

    $30\%$ &      $316$ &      $237$ &   $74,892$ &      $720$ &      $110$ &     $0.86$ & $(\pm 0.04)$ &   $0.0022$ \\

    $40\%$ &      $365$ &      $274$ &  $100,010$ &      $740$ &       $30$ &     $0.99$ & $(\pm 0.05)$ &   $0.0041$ \\

    $50\%$ &      $408$ &      $306$ &  $124,848$ &      $860$ &       $20$ &     $1.22$ & $(\pm 0.05)$ &   $0.0021$ \\

    $60\%$ &      $447$ &      $335$ &  $149,745$ &      $970$ &       $20$ &     $1.58$ & $(\pm 0.07)$ &   $0.0038$ \\

    $70\%$ &      $482$ &      $362$ &  $174,484$ &      $950$ &       $10$ &     $1.85$ & $(\pm 0.08)$ &   $0.0038$ \\

    $80\%$ &      $516$ &      $387$ &  $199,692$ &     $1050$ &       $10$ &     $2.13$ & $(\pm 0.10)$ &   $0.0018$ \\

    $90\%$ &      $547$ &      $410$ &  $224,270$ &     $1090$ &       $10$ &     $2.51$ & $(\pm 0.13)$ &   $0.0036$ \\

   $100\%$ &      $576$ &      $432$ &  $248,832$ &      $980$ &       $10$ &     $2.65$ & $(\pm 0.16)$ &   $0.0017$ \\
\hline
\end{tabular}
}
\label{tab:ComplexRalphKfix}
\end{table}

\begin{table}
\caption{Amount of iterations on each algorithm phase, execution times, and error rates on the ``Bird'' image with $10\%$ to $100\%$ of its original size and $k=10$. Each configuration is executed $100$ times to get the mean and standard deviation of the execution times.}
\resizebox{\textwidth}{!}{%
\begin{tabular}{ccccccccc}
\hline
{\bf Image Size} & {\bf Width} & {\bf Height} & {\bf Tot. Pixels} & {\bf Ph. 1} & {\bf Ph. 2} & \multicolumn{ 2}{c}{{\bf Time (s)}} & {\bf Error Rate} \\
\hline
    $10\%$ &      $253$ &      $190$ &   $48,070$ &      $870$ &       $10$ &     $0.49$ & $(\pm 0.07)$ &   $0.0031$ \\

    $20\%$ &      $358$ &      $269$ &   $96,302$ &      $950$ &       $10$ &     $1.04$ & $(\pm 0.13)$ &   $0.0012$ \\

    $30\%$ &      $439$ &      $329$ &  $144,431$ &     $1030$ &       $10$ &     $1.67$ & $(\pm 0.23)$ &   $0.0008$ \\

    $40\%$ &      $506$ &      $380$ &  $192,280$ &     $1080$ &       $10$ &     $2.28$ & $(\pm 0.34)$ &   $0.0004$ \\

    $50\%$ &      $566$ &      $425$ &  $240,550$ &     $1230$ &       $10$ &     $3.04$ & $(\pm 0.40)$ &   $0.0003$ \\

    $60\%$ &      $620$ &      $465$ &  $288,300$ &     $1240$ &       $10$ &     $3.96$ & $(\pm 0.67)$ &   $0.0004$ \\

    $70\%$ &      $670$ &      $502$ &  $336,340$ &     $1280$ &       $10$ &     $5.30$ & $(\pm 0.69)$ &   $0.0005$ \\

    $80\%$ &      $716$ &      $537$ &  $384,492$ &     $1530$ &       $10$ &     $6.70$ & $(\pm 1.00)$ &   $0.0002$ \\

    $90\%$ &      $759$ &      $570$ &  $432,630$ &     $1430$ &       $10$ &     $8.38$ & $(\pm 1.12)$ &   $0.0004$ \\

   $100\%$ &      $800$ &      $600$ &  $480,000$ &     $1500$ &       $10$ &    $10.26$ & $(\pm 1.32)$ &   $0.0002$ \\
\hline
\end{tabular}
}
\label{tab:ComplexBirdKfix}
\end{table}

\begin{figure}
\centering
\setlength\tabcolsep{1pt}
\begin{tabular}{cc}
    \subfloat{\includegraphics[height=3.8cm]{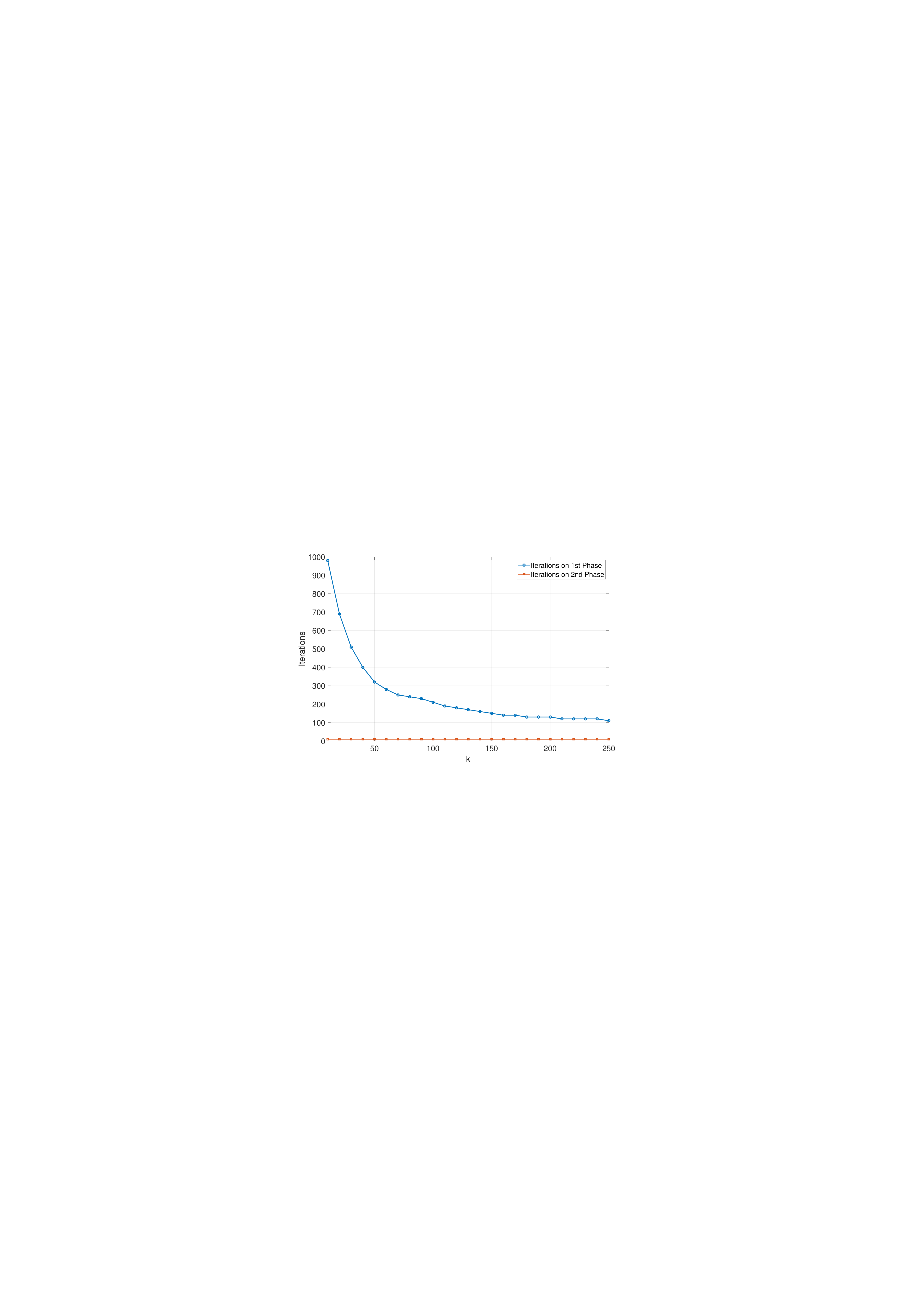}                    } &
    \subfloat{\includegraphics[height=3.8cm]{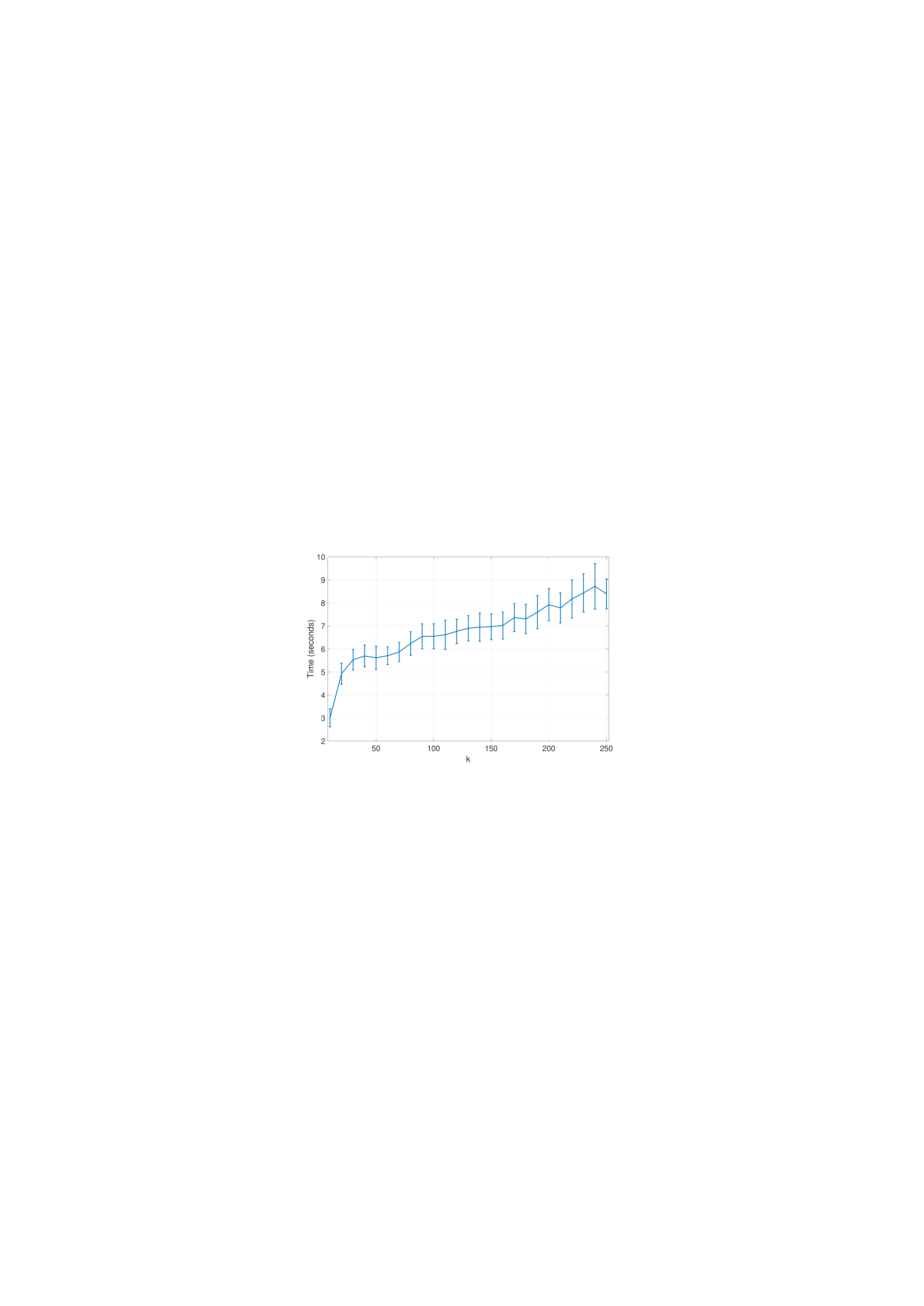}                    } \\
    \multicolumn{2}{c}{(b)} \\
    \subfloat{\includegraphics[height=3.8cm]{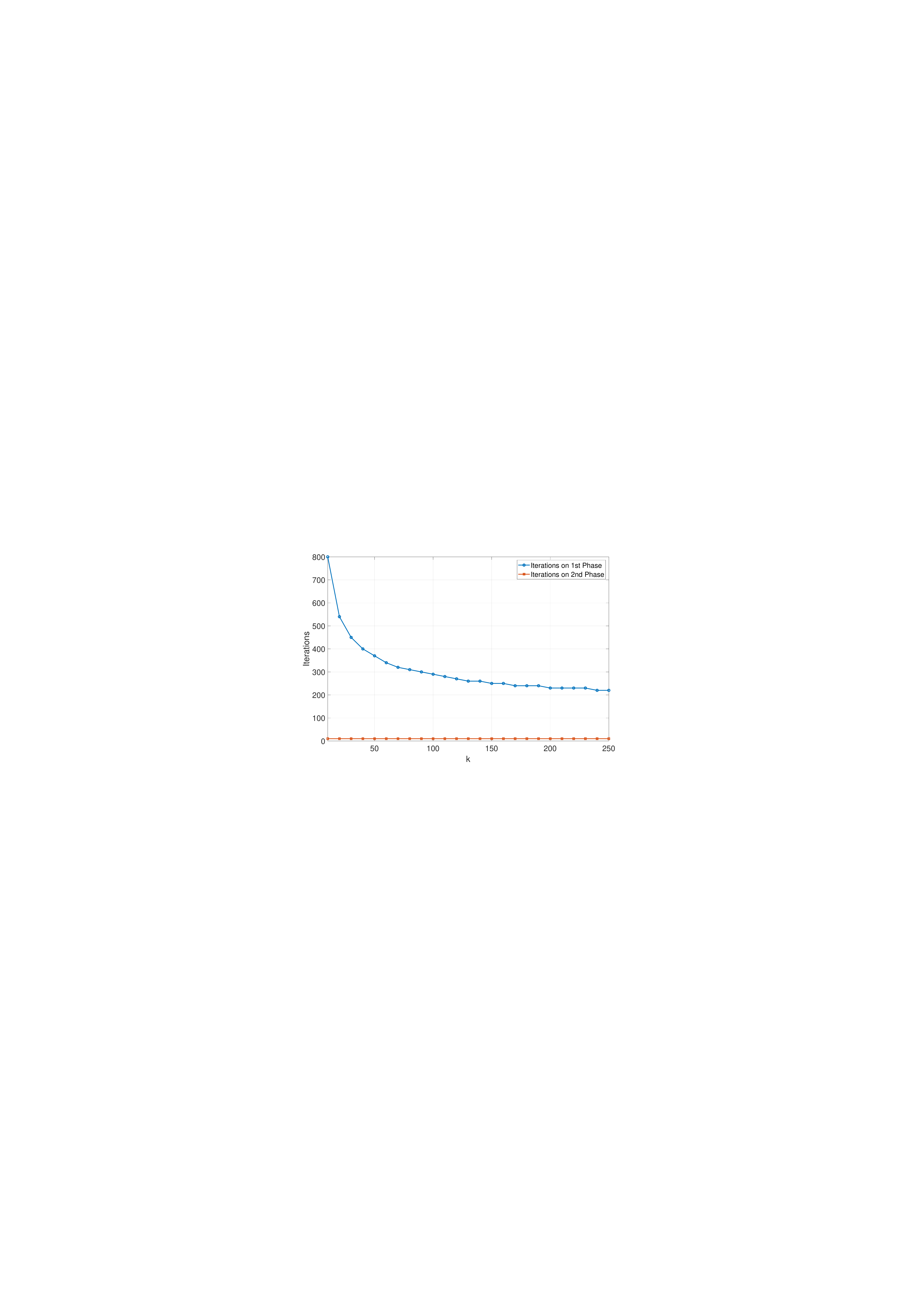}                    } &
    \subfloat{\includegraphics[height=3.8cm]{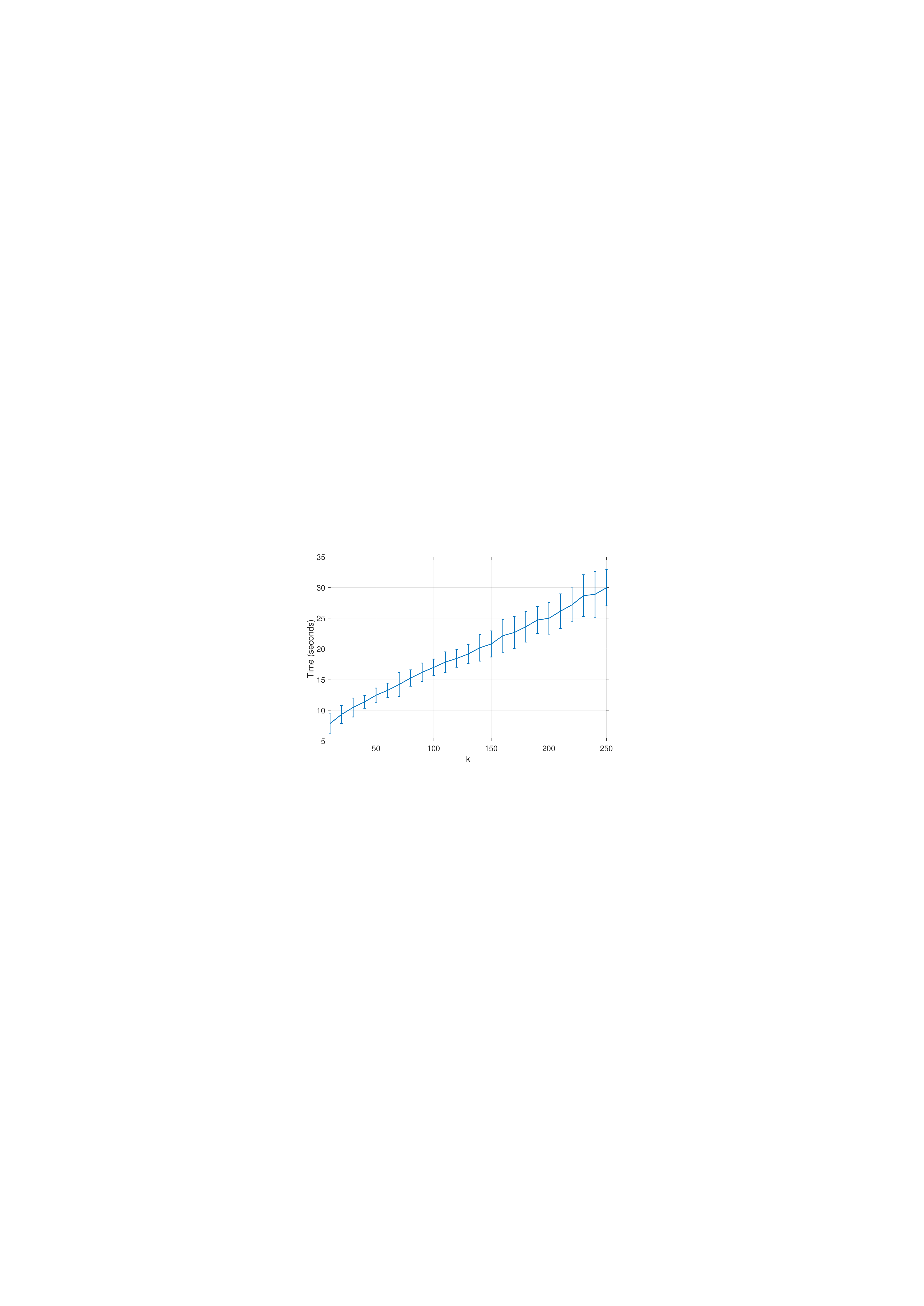}                    } \\
    \multicolumn{2}{c}{(a)} \\
\end{tabular}
\caption{Number of iterations (left) and time (right) required to the convergence, with images in their original size and $k=10$ to $k=250$. (a) ``Dog'' image. (b) ``Bird'' image. Each point in time traces is the average of $100$ realizations. The error bars represent standard deviation.}
\label{fig:complexkvar}
\end{figure}

Figure \ref{fig:complexkvar} and Tables \ref{tab:ComplexRalphKvar} and \ref{tab:ComplexBirdKvar} show the number of iterations of the outer loops of the first and second stages, and time required to convergence when the proposed method is applied to the same two images from Figure \ref{fig:TwoClass}: ``Dog'' and ``Bird''. However, this time the images are not resized and the out-degree of the nodes has increasing sizes ($k = \{10, 20, \ldots, 250\}$). By analyzing the graphics, it is possible to realize that as $k$ increases, the number of iterations of the first stage outer loop decreases. This is expected, since the network connectivity is increasing, and thus the labels have higher spread at each iteration. On the other hand, the execution time increases because the first stage inner loop execution time is higher as $k$ increases. However, this increase is only logarithmic ($O(\log n)$). Tables \ref{tab:ComplexRalphKvar} and \ref{tab:ComplexBirdKvar} also show the error rate in each scenario, which is calculated as previously described.

\begin{table}
\caption{Amount of iterations on each algorithm phase, execution times, and error rates on the ``Dog'' image with its original size and $k=10$ to $k=250$. Each configuration is executed $100$ times to get the mean and standard deviation of the execution times.}
\centering
\begin{tabular}{cccccc}
\hline
   {\bf k} & {\bf Ph. 1} & {\bf Ph. 2} & \multicolumn{ 2}{c}{{\bf Time (s)}} & {\bf Error Rate} \\
\hline
      $10$ &      $980$ &       $10$ &     $3.00$ & $(\pm 0.39)$ &   $0.0017$ \\

      $20$ &      $690$ &       $10$ &     $4.92$ & $(\pm 0.46)$ &   $0.0012$ \\

      $30$ &      $510$ &       $10$ &     $5.53$ & $(\pm 0.44)$ &   $0.0006$ \\

      $40$ &      $400$ &       $10$ &     $5.69$ & $(\pm 0.47)$ &   $0.0006$ \\

      $50$ &      $320$ &       $10$ &     $5.61$ & $(\pm 0.51)$ &   $0.0006$ \\

      $60$ &      $280$ &       $10$ &     $5.71$ & $(\pm 0.38)$ &   $0.0006$ \\

      $70$ &      $250$ &       $10$ &     $5.87$ & $(\pm 0.41)$ &   $0.0005$ \\

      $80$ &      $240$ &       $10$ &     $6.23$ & $(\pm 0.51)$ &   $0.0005$ \\

      $90$ &      $230$ &       $10$ &     $6.54$ & $(\pm 0.55)$ &   $0.0005$ \\

     $100$ &      $210$ &       $10$ &     $6.55$ & $(\pm 0.54)$ &   $0.0005$ \\

     $110$ &      $190$ &       $10$ &     $6.61$ & $(\pm 0.63)$ &   $0.0005$ \\

     $120$ &      $180$ &       $10$ &     $6.76$ & $(\pm 0.53)$ &   $0.0005$ \\

     $130$ &      $170$ &       $10$ &     $6.90$ & $(\pm 0.55)$ &   $0.0005$ \\

     $140$ &      $160$ &       $10$ &     $6.95$ & $(\pm 0.62)$ &   $0.0005$ \\

     $150$ &      $150$ &       $10$ &     $6.96$ & $(\pm 0.56)$ &   $0.0005$ \\

     $160$ &      $140$ &       $10$ &     $7.02$ & $(\pm 0.59)$ &   $0.0005$ \\

     $170$ &      $140$ &       $10$ &     $7.36$ & $(\pm 0.61)$ &   $0.0005$ \\

     $180$ &      $130$ &       $10$ &     $7.30$ & $(\pm 0.64)$ &   $0.0005$ \\

     $190$ &      $130$ &       $10$ &     $7.60$ & $(\pm 0.72)$ &   $0.0005$ \\

     $200$ &      $130$ &       $10$ &     $7.92$ & $(\pm 0.70)$ &   $0.0005$ \\

     $210$ &      $120$ &       $10$ &     $7.78$ & $(\pm 0.65)$ &   $0.0005$ \\

     $220$ &      $120$ &       $10$ &     $8.17$ & $(\pm 0.83)$ &   $0.0005$ \\

     $230$ &      $120$ &       $10$ &     $8.43$ & $(\pm 0.83)$ &   $0.0005$ \\

     $240$ &      $120$ &       $10$ &     $8.72$ & $(\pm 0.99)$ &   $0.0005$ \\

     $250$ &      $110$ &       $10$ &     $8.39$ & $(\pm 0.65)$ &   $0.0005$ \\
\hline
\end{tabular}

\label{tab:ComplexRalphKvar}
\end{table}

\begin{table}
\caption{Amount of iterations on each algorithm phase, execution times, and error rates on the ``Bird'' image with its original size and $k=10$ to $k=250$. Each configuration is executed $100$ times to get the mean and standard deviation of the execution times.}
\centering
\begin{tabular}{cccccc}
\hline
   {\bf k} & {\bf Ph. 1} & {\bf Ph. 2} & \multicolumn{ 2}{c}{{\bf Time (s)}} & {\bf Error Rate} \\
\hline
      $10$ &      $800$ &       $10$ &     $7.84$ & $(\pm 1.57)$ &   $0.0023$ \\

      $20$ &      $540$ &       $10$ &     $9.32$ & $(\pm 1.44)$ &   $0.0014$ \\

      $30$ &      $450$ &       $10$ &    $10.46$ & $(\pm 1.54)$ &   $0.0015$ \\

      $40$ &      $400$ &       $10$ &    $11.37$ & $(\pm 1.04)$ &   $0.0016$ \\

      $50$ &      $370$ &       $10$ &    $12.46$ & $(\pm 1.15)$ &   $0.0017$ \\

      $60$ &      $340$ &       $10$ &    $13.25$ & $(\pm 1.20)$ &   $0.0020$ \\

      $70$ &      $320$ &       $10$ &    $14.21$ & $(\pm 1.95)$ &   $0.0022$ \\

      $80$ &      $310$ &       $10$ &    $15.26$ & $(\pm 1.32)$ &   $0.0023$ \\

      $90$ &      $300$ &       $10$ &    $16.19$ & $(\pm 1.52)$ &   $0.0024$ \\

     $100$ &      $290$ &       $10$ &    $16.99$ & $(\pm 1.36)$ &   $0.0024$ \\

     $110$ &      $280$ &       $10$ &    $17.84$ & $(\pm 1.67)$ &   $0.0025$ \\

     $120$ &      $270$ &       $10$ &    $18.45$ & $(\pm 1.43)$ &   $0.0025$ \\

     $130$ &      $260$ &       $10$ &    $19.18$ & $(\pm 1.54)$ &   $0.0025$ \\

     $140$ &      $260$ &       $10$ &    $20.19$ & $(\pm 2.17)$ &   $0.0025$ \\

     $150$ &      $250$ &       $10$ &    $20.81$ & $(\pm 2.11)$ &   $0.0026$ \\

     $160$ &      $250$ &       $10$ &    $22.15$ & $(\pm 2.68)$ &   $0.0026$ \\

     $170$ &      $240$ &       $10$ &    $22.67$ & $(\pm 2.64)$ &   $0.0026$ \\

     $180$ &      $240$ &       $10$ &    $23.62$ & $(\pm 2.48)$ &   $0.0026$ \\

     $190$ &      $240$ &       $10$ &    $24.71$ & $(\pm 2.18)$ &   $0.0026$ \\

     $200$ &      $230$ &       $10$ &    $24.99$ & $(\pm 2.57)$ &   $0.0026$ \\

     $210$ &      $230$ &       $10$ &    $26.15$ & $(\pm 2.80)$ &   $0.0026$ \\

     $220$ &      $230$ &       $10$ &    $27.18$ & $(\pm 2.76)$ &   $0.0026$ \\

     $230$ &      $230$ &       $10$ &    $28.68$ & $(\pm 3.39)$ &   $0.0026$ \\

     $240$ &      $220$ &       $10$ &    $28.89$ & $(\pm 3.72)$ &   $0.0026$ \\

     $250$ &      $220$ &       $10$ &    $29.97$ & $(\pm 2.99)$ &   $0.0027$ \\
\hline
\end{tabular}

\label{tab:ComplexBirdKvar}
\end{table}

It is worth noting that in real-world problems, $c$ and $k$ do not increase proportionally to $n$ as image sizes increases. The amount of classes $c$ is unrelated to the image size, and the optimal value of $k$ depends on many factors, like the image structure, labeled pixels, objects position, and others, but even in similar images it does not need to increase linearly on $n$ to keep similar network connectivity, as the small-world property of complex networks applies. Therefore, the average time complexity in the first stage is usually lower than $O(n \log n)$.

In this sense, step 2 would dominate the execution time. However, although step 2 has the highest computational complexity, its execution time is faster than the first stage loop in all the experiments presented in this paper. It will undoubtedly dominate the execution time for huge images, but in the typical scenario, the execution time is still dominated by the first stage (steps 5 to 8).

The second stage execution time is usually negligible. It is very fast when compared to the first stage and step 2.

In summary, steps 5 to 8 run at linear time $O(n)$ in the best scenario (fixed $k$) and linearithmic time $O(n \log n)$ in the worst scenario. So, the first stage dominates the execution time in images of moderate size. Only in huge images, step 2 would dominate the execution time, and it runs in linearithmic time $O(n \log n)$. Therefore, in most real-world scenarios a time complexity from $O(n)$ to $O(n \log n)$ is expected.

\subsection{Storage Complexity}
\label{sec:StorageComplexity}

Regarding the memory requirements and storage complexity, the proposed algorithm uses the following data structures: the resized image, the features table, the neighbors table, the weights table, the domination vectors, and the labeled output image. The resized image size is $\frac{n}{9}$. The features table is built from the input image (or the resized image), and it is used to build the graph. There are $9$ features, so the features table size is $n$ in the first stage (which works with the resized image) and $9n$ on the second stage. In the first stage, the neighbors table holds the $k$-nearest neighbors of each node, so its size is $\frac{kn}{9}$. In the second stage, each neighbor has only $8$ neighbors or less, so the neighbors table size is $8n$. The weights table has the same size as the neighbors table in both stages. The domination vectors hold the pertinence of each node to each class, so its size is $\frac{cn}{9}$ in the first stage, and $cn$ in the second stage. Finally, the labeled output image size is $n$. As explained before, in real-world problems, $c$ and $k$ do not increase proportionally to $n$. So, we may expect that all data structures grow linearly on $n$ and the storage complexity is $O(n)$.

\subsection{Large-Scale Networks}

The proposed method was also tested on large images to evaluate its behavior on large-scale networks. The source picture of the ``Dog'' image from Figure \ref{fig:TwoClass} is used in these experiments. It is a $16$ megapixels JPEG image ($4608 \times 3456$ pixels).

\begin{figure}
\centering
\setlength\tabcolsep{1pt}
\begin{tabular}{cc}
    \subfloat[]{\includegraphics[height=3.8cm]{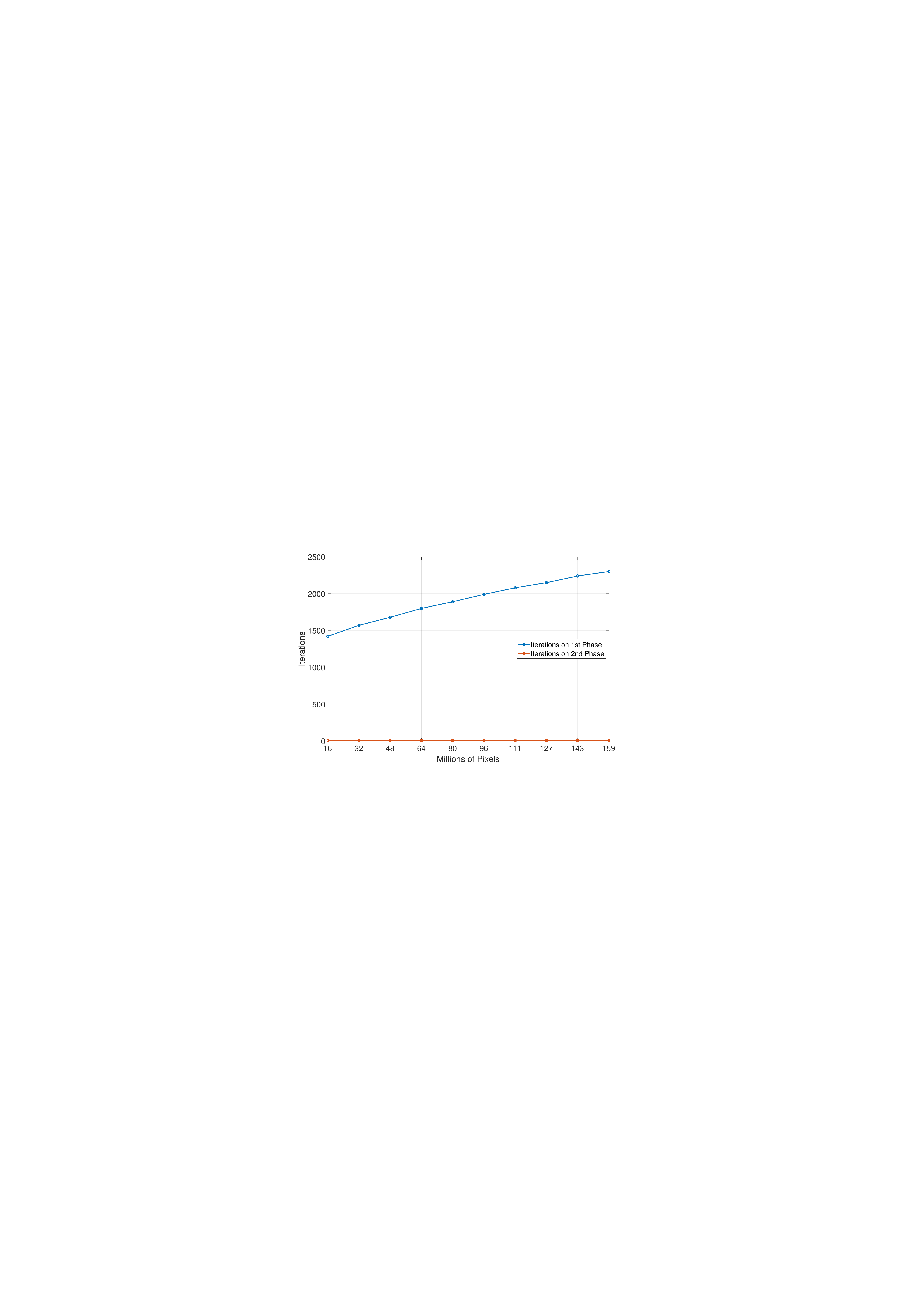}                    } &
    \subfloat[]{\includegraphics[height=3.8cm]{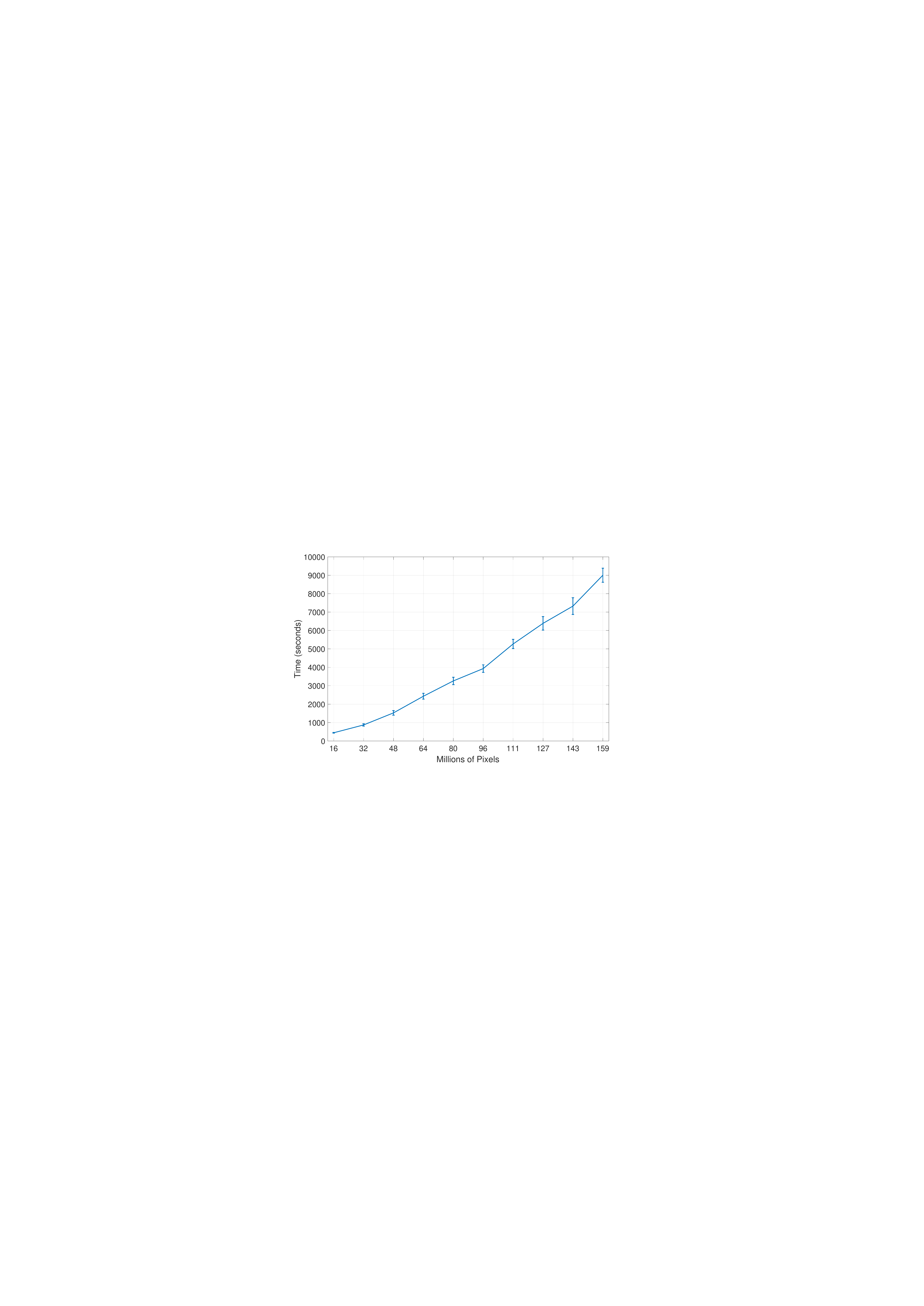}                    } \\
\end{tabular}
\caption{(a) Number of iterations and (b) time required to the convergence with the ``Dog'' image with $16$ megapixels to $159$ megapixels and $k=10$. Each point in (b) is the average of $7$ realizations. The error bars represent standard deviation.}
\label{fig:largescalekfix}
\end{figure}

In the first experiment, the picture is resized up to $10$ times its size, using bicubic interpolation, to simulate a $159$ megapixels picture. After the enlargement, some Poisson noise is added using the ``imnoise'' function from the MATLAB Image Processing Toolbox to simulate the noise from a camera sensor. Otherwise, the enlarged images would look like a set of flat tiles. The same enlargement is applied to the ``scribbles'' image, but using the nearest-neighbor interpolation to avoid the introduction of new colors which would be mistakenly interpreted as new classes. Figure \ref{fig:largescalekfix} and Table \ref{tab:LargeScaleRalphKfix} show the number of iterations of the outer loops of the first and second stages, and time required to convergence when the proposed method is applied. Table \ref{tab:LargeScaleRalphKfix} additionally shows the error rate in each scenario.

\begin{table}
\caption{Amount of iterations on each algorithm phase, execution times, and error rates on the ``Dog'' image with $16$ megapixels to $159$ megapixels and $k=10$. Each configuration is executed $7$ times to get the mean and standard deviation of the execution times.}
\resizebox{\textwidth}{!}{%
\begin{tabular}{ccccccccc}
\hline
{\bf Image Size} & {\bf Width} & {\bf Height} & {\bf Tot. Pixels} & {\bf Ph. 1} & {\bf Ph. 2} & \multicolumn{ 2}{c}{{\bf Time (s)}} & {\bf Error Rate} \\
\hline
   $16$ MP &     $4608$ &     $3456$ & $15,925,248$ &     $1420$ &       $10$ &   $445.38$ & $(\pm 16.21)$ &   $0.0132$ \\

   $32$ MP &     $6517$ &     $4888$ & $31,855,096$ &     $1570$ &       $10$ &   $870.52$ & $(\pm 64.50)$ &   $0.0140$ \\

   $48$ MP &     $7982$ &     $5986$ & $47,780,252$ &     $1680$ &       $10$ & $1,526.19$ & $(\pm 126.27)$ &   $0.0140$ \\

   $64$ MP &     $9216$ &     $6912$ & $63,700,992$ &     $1800$ &       $10$ & $2,430.29$ & $(\pm 159.01)$ &   $0.0134$ \\

   $80$ MP &    $10304$ &     $7728$ & $79,629,312$ &     $1890$ &       $10$ & $3,261.71$ & $(\pm 201.10)$ &   $0.0135$ \\

   $95$ MP &    $11288$ &     $8466$ & $95,564,208$ &     $1990$ &       $10$ & $3,931.04$ & $(\pm 205.11)$ &   $0.0137$ \\

  $111$ MP &    $12192$ &     $9144$ & $111,483,648$ &     $2080$ &       $10$ & $5,268.22$ & $(\pm 252.83)$ &   $0.0138$ \\

  $127$ MP &    $13034$ &     $9776$ & $127,420,384$ &     $2150$ &       $10$ & $6,387.76$ & $(\pm 365.23)$ &   $0.0154$ \\

  $143$ MP &    $13824$ &    $10368$ & $143,327,232$ &     $2240$ &       $10$ & $7,324.84$ & $(\pm 454.97)$ &   $0.0157$ \\

  $159$ MP &    $14572$ &    $10929$ & $159,257,388$ &     $2300$ &       $10$ & $9,001.67$ & $(\pm 380.48)$ &   $0.0159$ \\
\hline
\end{tabular}
}
\label{tab:LargeScaleRalphKfix}
\end{table}

\begin{figure}
\centering
\setlength\tabcolsep{1pt}
\begin{tabular}{cc}
    \subfloat[]{\includegraphics[height=3.8cm]{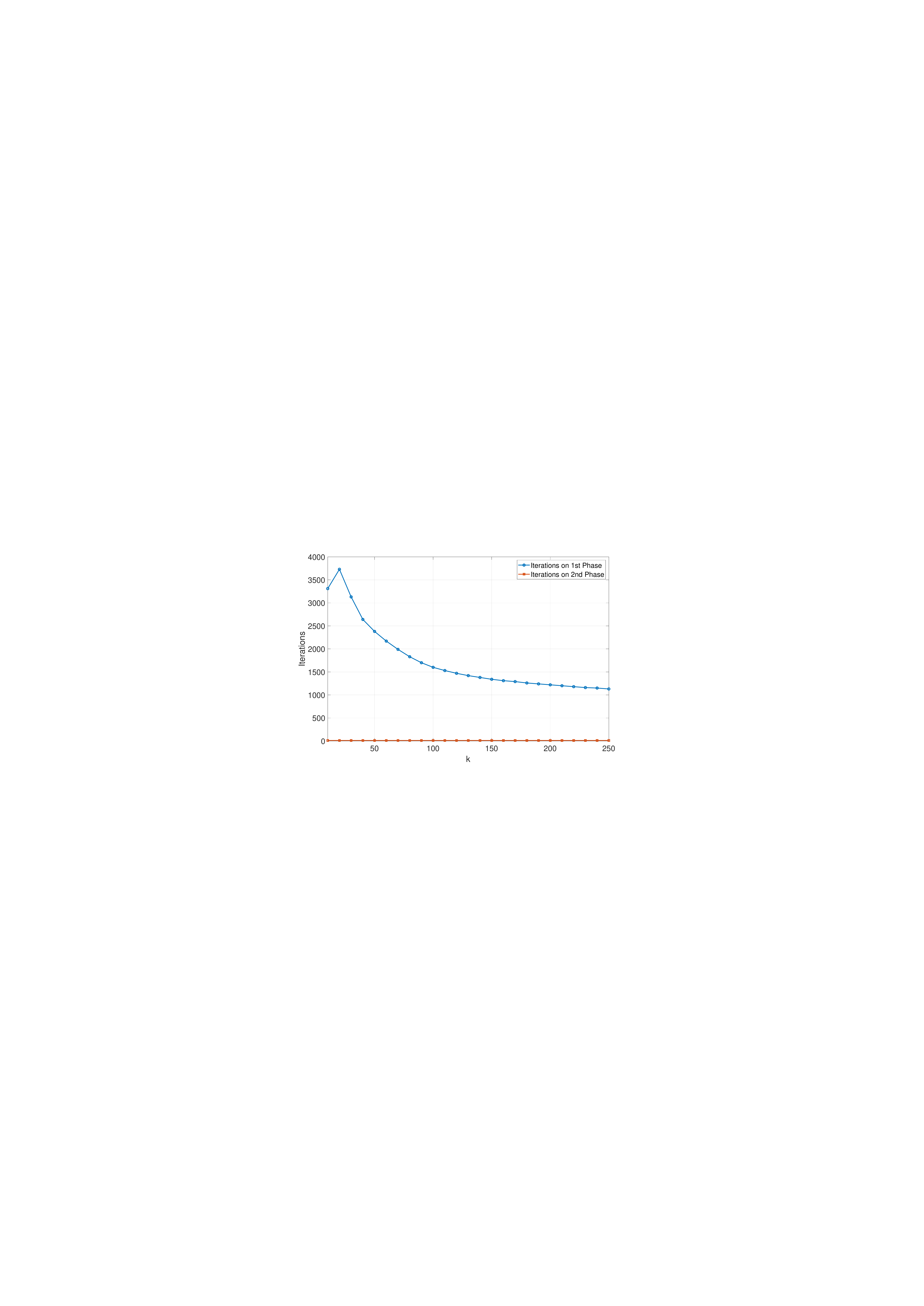}                    } &
    \subfloat[]{\includegraphics[height=3.8cm]{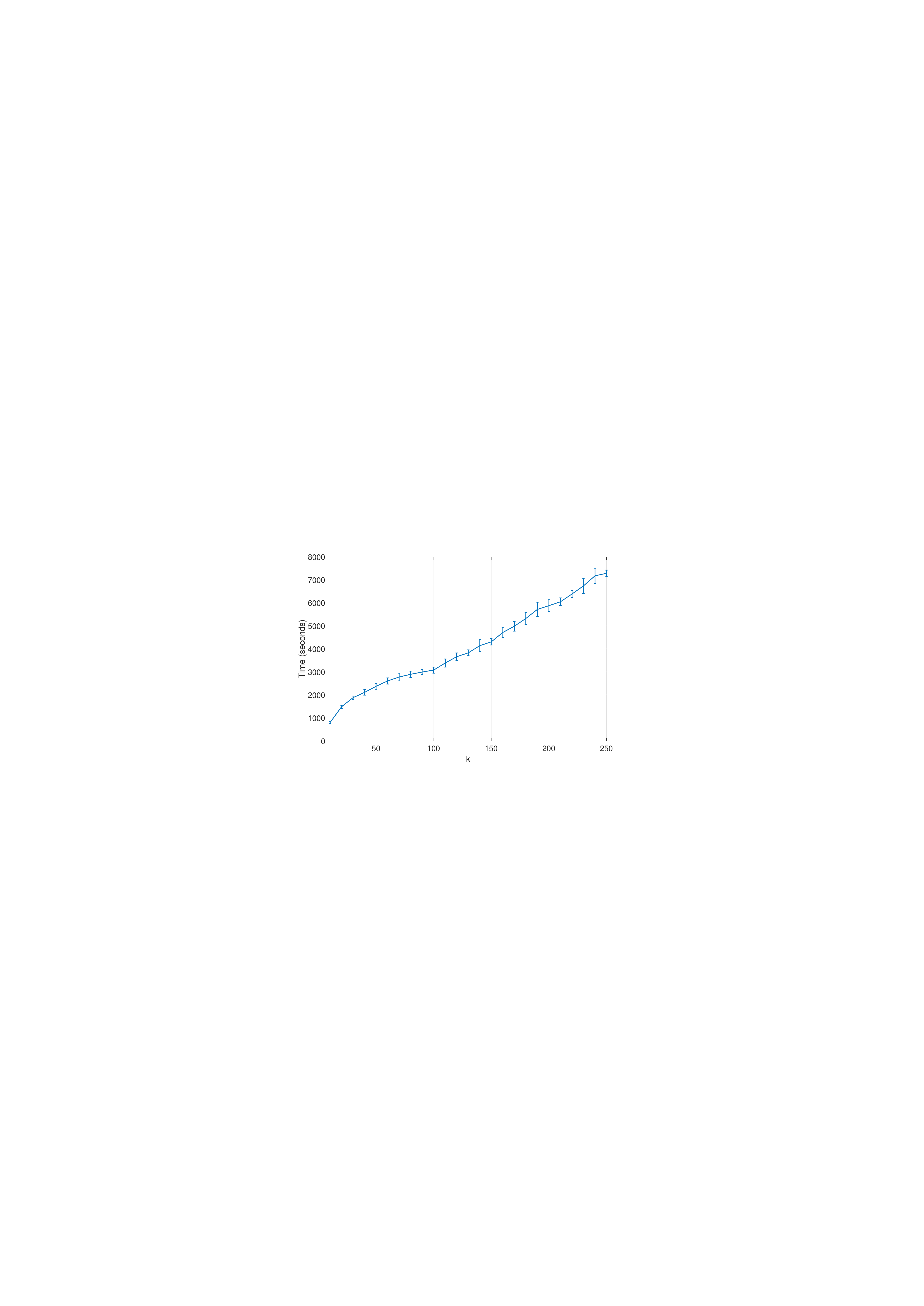}                    } \\
\end{tabular}
\caption{(a) Number of iterations and (b) time required to the convergence with the ``Dog'' image with $16$ megapixels and $k=10$ to $k=250$. Each point in (b) is the average of $10$ realizations. The error bars represent standard deviation.}
\label{fig:largescalekvar}
\end{figure}

In the second experiment, the $16$ megapixels picture is used without modification, but the out-degree of the nodes has increasing sizes ($k = \{10, 20, \ldots, 250\}$). Figure \ref{fig:largescalekvar} and Table \ref{tab:LargeScaleRalphKvar} show the number of iterations of the outer loops of the first and second stages, and time required to convergence when the proposed method is applied. Table \ref{tab:LargeScaleRalphKvar} additionally shows the error rate in each scenario.

\begin{table}
\caption{Amount of iterations on each algorithm phase, execution times, and error rates on the ``Dog'' image with $16$ megapixels and $k=10$ to $k=250$. Each configuration is executed $9$ times to get the mean and standard deviation of the execution times.}
\centering
\begin{tabular}{cccccc}
\hline
   {\bf k} & {\bf Ph. 1} & {\bf Ph. 2} & \multicolumn{ 2}{c}{{\bf Time (s)}} & {\bf Error Rate} \\
\hline
      $10$ &     $3310$ &       $10$ &   $797.06$ & $(\pm 45.72)$ &   $0.0098$ \\

      $20$ &     $3730$ &       $10$ & $1,485.54$ & $(\pm 71.60)$ &   $0.0115$ \\

      $30$ &     $3130$ &       $10$ & $1,882.20$ & $(\pm 70.04)$ &   $0.0120$ \\

      $40$ &     $2640$ &       $10$ & $2,110.70$ & $(\pm 116.52)$ &   $0.0123$ \\

      $50$ &     $2380$ &       $10$ & $2,376.00$ & $(\pm 127.28)$ &   $0.0125$ \\

      $60$ &     $2170$ &       $10$ & $2,607.85$ & $(\pm 133.23)$ &   $0.0125$ \\

      $70$ &     $1990$ &       $10$ & $2,779.26$ & $(\pm 170.00)$ &   $0.0126$ \\

      $80$ &     $1830$ &       $10$ & $2,897.11$ & $(\pm 141.70)$ &   $0.0127$ \\

      $90$ &     $1700$ &       $10$ & $2,994.93$ & $(\pm 107.31)$ &   $0.0126$ \\

     $100$ &     $1600$ &       $10$ & $3,081.27$ & $(\pm 131.31)$ &   $0.0126$ \\

     $110$ &     $1530$ &       $10$ & $3,388.25$ & $(\pm 172.80)$ &   $0.0127$ \\

     $120$ &     $1470$ &       $10$ & $3,662.71$ & $(\pm 161.67)$ &   $0.0127$ \\

     $130$ &     $1420$ &       $10$ & $3,828.58$ & $(\pm 126.99)$ &   $0.0128$ \\

     $140$ &     $1380$ &       $10$ & $4,141.11$ & $(\pm 256.80)$ &   $0.0128$ \\

     $150$ &     $1340$ &       $10$ & $4,307.77$ & $(\pm 140.86)$ &   $0.0129$ \\

     $160$ &     $1310$ &       $10$ & $4,715.17$ & $(\pm 229.51)$ &   $0.0129$ \\

     $170$ &     $1290$ &       $10$ & $4,986.44$ & $(\pm 210.99)$ &   $0.0130$ \\

     $180$ &     $1260$ &       $10$ & $5,323.27$ & $(\pm 262.29)$ &   $0.0130$ \\

     $190$ &     $1240$ &       $10$ & $5,717.48$ & $(\pm 317.35)$ &   $0.0131$ \\

     $200$ &     $1220$ &       $10$ & $5,878.81$ & $(\pm 257.88)$ &   $0.0118$ \\

     $210$ &     $1200$ &       $10$ & $6,047.86$ & $(\pm 166.69)$ &   $0.0113$ \\

     $220$ &     $1180$ &       $10$ & $6,385.12$ & $(\pm 142.22)$ &   $0.0113$ \\

     $230$ &     $1160$ &       $10$ & $6,739.01$ & $(\pm 333.90)$ &   $0.0112$ \\

     $240$ &     $1150$ &       $10$ & $7,176.41$ & $(\pm 328.65)$ &   $0.0112$ \\

     $250$ &     $1130$ &       $10$ & $7,285.81$ & $(\pm 138.61)$ &   $0.0113$ \\
\hline
\end{tabular}
\label{tab:LargeScaleRalphKvar}
\end{table}

By analyzing these results, the same patterns seem on the experiments with smaller images is observed. As the network increases, the amount of first phase iterations also increases and the execution time increases almost linearly. As the connectivity increases, the amount of first phase iterations decreases and the execution time increases logarithmically.

\subsection{Small-World-Ness}

The proposed method efficiency highly relies on the \emph{small-world} property of the networks it generates in its first phase. In particular, when the edges are created to the $k$-nearest neighbors of each node, the clustering coefficient is usually high, so the label information is quickly spread to the neighborhood.

To verify if the \emph{small-world} property is present on a network, \citet{Humphries2008} proposed a measure called \emph{small-world-ness}. The \emph{small-world-ness} $S$ of a given graph may be calculated as follows:
\begin{equation}
S = \frac{C}{C_{\textrm{rand}}} \frac{L_{\textrm{rand}}}{L}
\end{equation}
where $C$ and $L$ are the clustering coefficient and the mean shortest path length of the network, respectively, and $C_{\textrm{rand}}$ and $L_{\textrm{rand}}$ are the clustering coefficient and the mean shortest path length observed in random equivalente networks, i.e., networks with the same amount of nodes and edges. $S>1$ indicates the presence of the small-world property.

Unfortunately, $S$ is undefined for disconnected networks, because in those scenarios $L$ diverges to infinity. To overcome this drawback, \citet{Zanin2015} proposed an alternative formulation to compute \emph{small-world-ness}, which uses the average efficiency of the network instead of the shortest path length since efficiency is defined even for disconnected networks. The efficiency $E$ of a graph $G$ is calculated as follows:

\begin{equation}
E(G) = \frac{1}{n(n-1)} \sum_{i \neq j \in G} \frac{1}{d_{ij}}
\end{equation}
where $n$ is the total of nodes in the network and $d_{ij}$ denotes the length of the shortest path between a node $i$ and another node $j$.

The new efficiency-based \emph{small-world-ness} $S^E$ is then defined as follows:
\begin{equation}
S^E = \frac{C}{C_{\textrm{rand}}} \frac{E}{E_{\textrm{rand}}}
\end{equation}
where $C$ and $E$ are the clustering coefficient and the average efficiency of the network, respectively, and $C_{\textrm{rand}}$ and $E_{\textrm{rand}}$ are the clustering coefficient and the average efficiency observed in random equivalent networks. $S^E>1$ indicates the presence of the small-world property.

Notice that disconnected networks are not a problem for the proposed method. An unlabeled node only needs a path to a labeled node to get label information. Even if an unlabeled node does not have a path to a labeled node, it still gets its label in the second stage. Therefore, the efficiency-based \emph{small-world-ness} is used in this paper.

Table \ref{tab:SmallWorldNessSize} shows the measures of \emph{small-world-ness}, clustering coefficient and efficiency for the networks built during the proposed method first phase for the ``Dog'' image with $10\%$ to $100\%$ of its original size and $k=10$. Table \ref{tab:SmallWorldNessK} shows the same measures for the networks built during the proposed method first phase for the ``Dog'' image with its original size and $k=10$ to $k=250$. In both cases, the mean clustering coefficient and the mean average efficiency of $20$ random networks with the same amount of nodes and edges are also shown for comparison.

By analysing Tables \ref{tab:SmallWorldNessSize} and \ref{tab:SmallWorldNessK}, one can notice that all the networks have high \emph{small-world-ness} levels, clearly showing that they have the small-world property. In particular, the clustering coefficients are much higher than those of an equivalent random network.

\begin{table}
\caption{Small-world-ness, clustering coefficient, and the average efficiency of the networks built during the first phase of the proposed method for the ``Dog'' image with $10\%$ to $100\%$ of its original size and $k=10$. The mean clustering coefficient and mean average efficiency of $20$ random network with the same amount of nodes and edges are also shown for comparison.}
\centering
\begin{tabular}{cccccc}
\hline
{\bf Image Size} & {\bf $S^E$} &  {\bf $C$} &  {\bf $E$} & {\bf $C_{\textrm{rand}}$} & {\bf $E_{\textrm{rand}}$} \\
\hline
      10\% &      38.03 &     0.4301 &     0.0898 &     0.0036 &     0.2816 \\

      20\% &      62.65 &     0.3923 &     0.0740 &     0.0018 &     0.2591 \\

      30\% &     127.09 &     0.3851 &     0.0685 &     0.0009 &     0.2234 \\

      40\% &     240.52 &     0.3739 &     0.0628 &     0.0005 &     0.1874 \\

      50\% &     418.88 &     0.3718 &     0.0603 &     0.0003 &     0.1610 \\

      60\% &     694.56 &     0.3683 &     0.0592 &     0.0002 &     0.1374 \\

      70\% &    1045.17 &     0.3648 &     0.0567 &     0.0002 &     0.1169 \\

      80\% &    1506.40 &     0.3625 &     0.0559 &     0.0001 &     0.0993 \\

      90\% &    2382.55 &     0.3606 &     0.0555 &     0.0001 &     0.0827 \\

     100\% &    3943.10 &     0.3595 &     0.0544 &     0.0001 &     0.0694 \\
\hline
\end{tabular}

\label{tab:SmallWorldNessSize}
\end{table}

\begin{table}
\caption{Small-world-ness, clustering coefficient, and the average efficiency of the networks built during the proposed method first phase for the ``Dog'' image with its original size and $k=10$ to $k=250$. The mean clustering coefficient and mean average efficiency of $20$ random network with the same amount of nodes and edges are also shown for comparison.}
\centering
\begin{tabular}{cccccc}
\hline
   {\bf k} & {\bf $S^E$} &  {\bf $C$} &  {\bf $E$} & {\bf $C_{\textrm{rand}}$} & {\bf $E_{\textrm{rand}}$} \\
\hline
        10 &    2599.98 &     0.3595 &     0.0544 &     0.0001 &     0.0695 \\

        20 &    4490.24 &     0.3858 &     0.0780 &     0.0001 &     0.0694 \\

        30 &    6446.55 &     0.3976 &     0.0935 &     0.0001 &     0.0694 \\

        40 &    6570.85 &     0.4061 &     0.1044 &     0.0001 &     0.0696 \\

        50 &    8705.99 &     0.4128 &     0.1134 &     0.0001 &     0.0696 \\

        60 &    8098.44 &     0.4178 &     0.1208 &     0.0001 &     0.0694 \\

        70 &    8161.99 &     0.4219 &     0.1272 &     0.0001 &     0.0695 \\

        80 &    7588.72 &     0.4255 &     0.1331 &     0.0001 &     0.0696 \\

        90 &   10167.96 &     0.4288 &     0.1385 &     0.0001 &     0.0695 \\

       100 &   10150.12 &     0.4319 &     0.1433 &     0.0001 &     0.0694 \\

       110 &    8386.42 &     0.4348 &     0.1475 &     0.0001 &     0.0694 \\

       120 &    9803.54 &     0.4375 &     0.1516 &     0.0001 &     0.0694 \\

       130 &   12145.72 &     0.4401 &     0.1556 &     0.0001 &     0.0695 \\

       140 &   10484.87 &     0.4426 &     0.1592 &     0.0001 &     0.0696 \\

       150 &   13139.53 &     0.4450 &     0.1627 &     0.0001 &     0.0696 \\

       160 &   12865.17 &     0.4471 &     0.1659 &     0.0001 &     0.0694 \\

       170 &   10555.00 &     0.4491 &     0.1689 &     0.0001 &     0.0695 \\

       180 &   11310.11 &     0.4510 &     0.1718 &     0.0001 &     0.0696 \\

       190 &   12677.64 &     0.4529 &     0.1747 &     0.0001 &     0.0695 \\

       200 &   14342.62 &     0.4547 &     0.1774 &     0.0001 &     0.0694 \\

       210 &   13331.34 &     0.4565 &     0.1801 &     0.0001 &     0.0694 \\

       220 &   15375.38 &     0.4583 &     0.1826 &     0.0001 &     0.0697 \\

       230 &   14912.54 &     0.4600 &     0.1851 &     0.0001 &     0.0696 \\

       240 &   15552.47 &     0.4618 &     0.1875 &     0.0001 &     0.0695 \\

       250 &   12742.60 &     0.4635 &     0.1899 &     0.0001 &     0.0694 \\
\hline
\end{tabular}

\label{tab:SmallWorldNessK}
\end{table}

\section{Benchmark}
\label{sec:Benchmark}

Figures \ref{fig:TwoClass} and \ref{fig:MultiClass} show segmentation examples on real-world images, where the user input is limited to a set of ``scribbles'' on the main object(s) and the background. The results are qualitatively good as they mostly agree with perceptual boundaries.

For quantitative results, the proposed method is applied to the $50$ images of the Microsoft GrabCut dataset \citep{Rother2004}. Though there are some other data sets available with ground truth segmentation results, this one is, to the best of my knowledge, the only one where seed regions are provided. It is also the only database which was widely used in other papers; therefore it is possible to present a quantitative comparison with state-of-the-art methods. Their original seed regions are not presented as ``scribbles''. Instead, they present a large number of labeled pixels and a narrow band around the contour of objects to be segmented. In spite of that, the proposed method can be applied to it without any modification or extra cost.

Table \ref{tab:GrabCutComparison} presents a comparison of the average error rates obtained on the GrabCut dataset \citep{Rother2004} by the proposed method and other interactive image segmentation methods. The proposed method was first applied to the whole dataset with its default parameters ($k=10$, $\mathbf{\lambda} = \mathbf{\lambda_1}$, $\omega=10^{-4}$). In this way, it achieved an error rate of $4.15\%$. Later, the parameter $k$ was optimized for each image, and an error rate of $3.21\%$ was achieved.

\begin{table}
\caption{Comparison of the average error rates obtained on the GrabCut dataset \citep{Rother2004} by the proposed method and other interactive image segmentation methods. The error rates for the other methods were compiled from the works of \citet{Ding2012}, \citet{Ducournau2014}, \citet{Wang2018b}, and \citet{Bampis2017}.}
\centering
\resizebox{\textwidth}{!}{%
\begin{tabular}{lc}
\hline
{\bf Method} & {\bf Error Rate} \\
\hline
sDPMNL (boundary) \citep{Ding2012} &    11.43\% \\

GMMVL (location + color + boundary) \citep{Yi2004} &    10.45\% \\

SVM (location + color + boundary) \citep{Chang2011} &     9.21\% \\

GM-MRF \citep{Blake2004} &     7.90\% \\

sDPMNL (color) \citep{Ding2012} &     7.65\% \\

Superpixels Hypergraph \citep{Ding2008} &     7.30\% \\

Lazy Snapping \citep{Li2004} &     6.65\% \\

Graph Cuts \citep{Boykov2001} &     6.60\% \\

Cost volume filtering \citep{Hosni2013} &     6.20\% \\

Directed Image Neighborhood Hypergraph \citep{Ducournau2014} &     6.15\% \\

RobustP\textsuperscript{n} \citep{Kohli2009} &     6.08\% \\

Grabcut \citep{Rother2004} &     5.46\% \\

Regularized Laplacian \citep{Duchenne2008} &     5.40\% \\

Grady's random walker \citep{Grady2006} &     5.40\% \\

Probabilistic Hypergraph \citep{Ding2010} &     5.33\% \\

DPMVL (color + boundary) \citep{Ding2012} &     5.19\% \\

Laplacian Coordinates \citep{Casaca2014} &     5.04\% \\

sDPMVL (color + boundary) \citep{Ding2012} &     4.78\% \\

Sub-Markov Random Walk \citep{Dong2016} &     4.61\% \\

Normalized Lazy Random Walker \citep{Bampis2017} &     4.37\% \\

Normalized Random Walker \citep{Bampis2017} &     4.35\% \\

Nonparametric Higher-Order \citep{Kim2010} &     4.25\% \\

{\bf Proposed Method (default parameters)} & {\bf 4.15\%} \\

Constrained Random Walks \citep{Yang2010} &     4.08\% \\

Lazy Randow Walks \citep{Shen2014} &     3.89\% \\

Robust Multilayer Graph Constraints \citep{Wang2016b} &     3.79\% \\

Texture Aware Model \citep{Zhou2013} &     3.64\% \\

Pairwise Likelihood Learning \citep{Wang2017} &     3.49\% \\

Multi-layer Graph Constraints \citep{Wang2016} &     3.44\% \\

{\bf Proposed Method (optimized $k$)} & {\bf 3.21\%} \\

Random Walks with Restart \citep{Kim2008} &     3.11\% \\

Normalized Sub-Markov Random Walk \citep{Bampis2017} &     3.10\% \\

Difusive Likelihood \citep{Wang2018b} &     3.08\% \\
\hline
\end{tabular}

}
\label{tab:GrabCutComparison}
\end{table}

\begin{figure}
\centering
\setlength\tabcolsep{1pt}
\begin{tabular}{cccc}
\subfloat{\includegraphics[width=2.8cm,height=2.3cm,keepaspectratio]{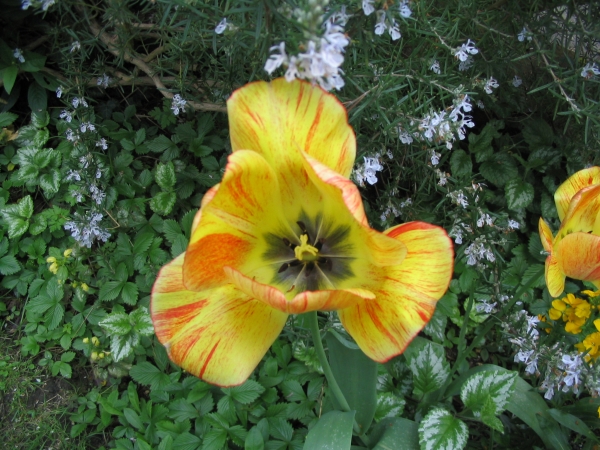}                    } &
\subfloat{\includegraphics[width=2.8cm,height=2.3cm,keepaspectratio]{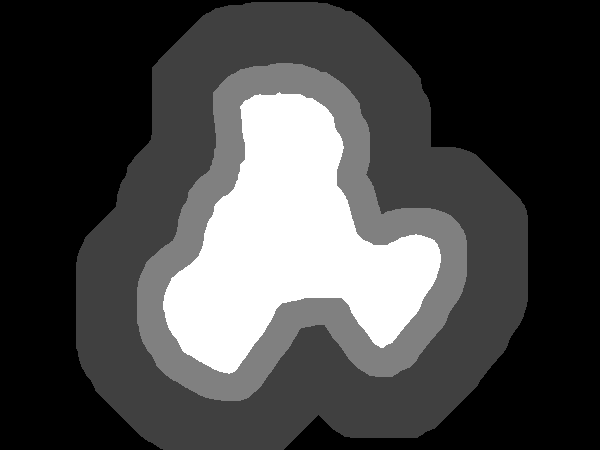}                    } &
\subfloat{\includegraphics[width=2.8cm,height=2.3cm,keepaspectratio]{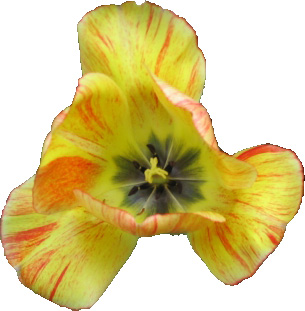}                    } &
\subfloat{\includegraphics[width=2.8cm,height=2.3cm,keepaspectratio]{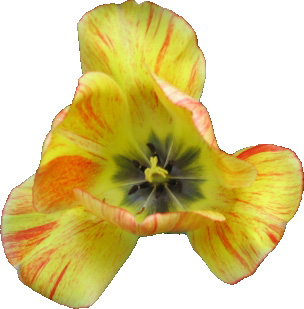}                    } \\
\subfloat{\includegraphics[width=2.8cm,height=2.3cm,keepaspectratio]{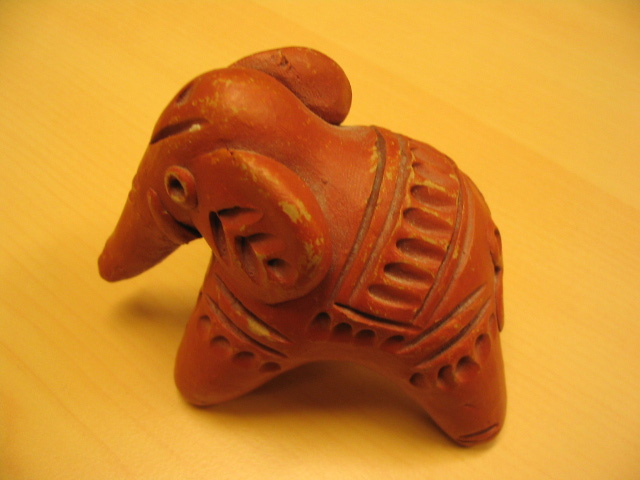}                    } &
\subfloat{\includegraphics[width=2.8cm,height=2.3cm,keepaspectratio]{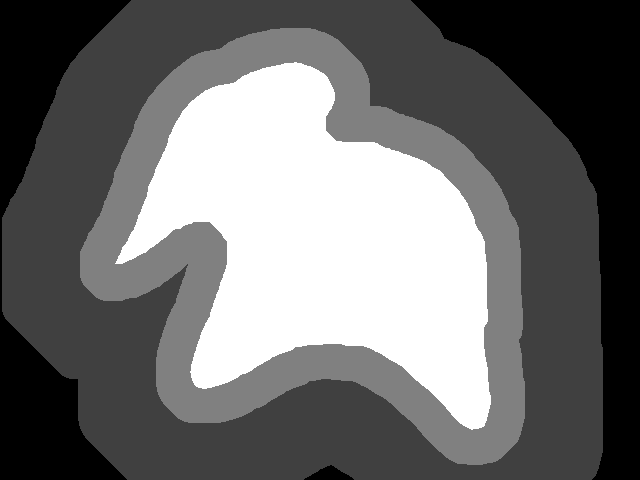}                    } &
\subfloat{\includegraphics[width=2.8cm,height=2.3cm,keepaspectratio]{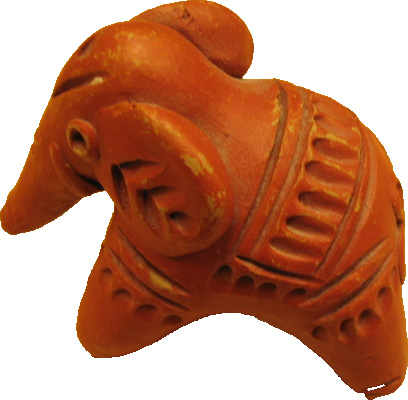}                    } &
\subfloat{\includegraphics[width=2.8cm,height=2.3cm,keepaspectratio]{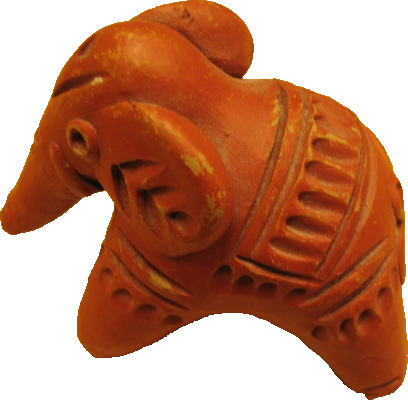}                    } \\
\subfloat{\includegraphics[width=2.8cm,height=2.3cm,keepaspectratio]{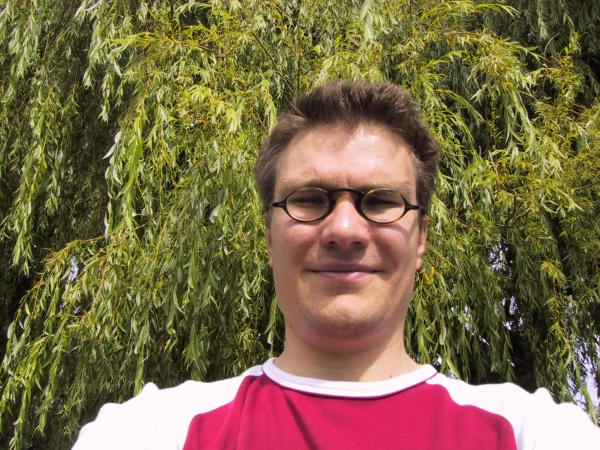}                    } &
\subfloat{\includegraphics[width=2.8cm,height=2.3cm,keepaspectratio]{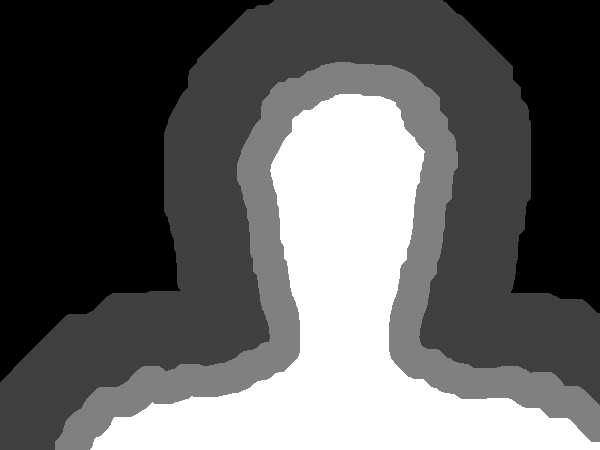}                    } &
\subfloat{\includegraphics[width=2.8cm,height=2.3cm,keepaspectratio]{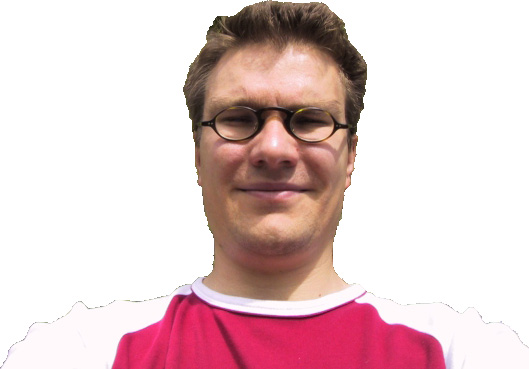}                    } &
\subfloat{\includegraphics[width=2.8cm,height=2.3cm,keepaspectratio]{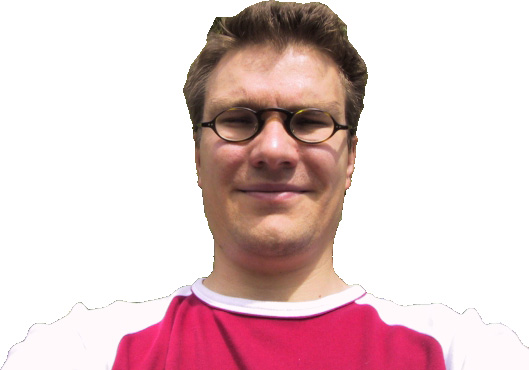}                    } \\
\subfloat{\includegraphics[width=2.8cm,height=2.3cm,keepaspectratio]{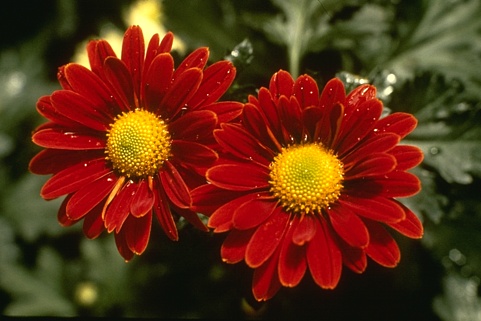}                    } &
\subfloat{\includegraphics[width=2.8cm,height=2.3cm,keepaspectratio]{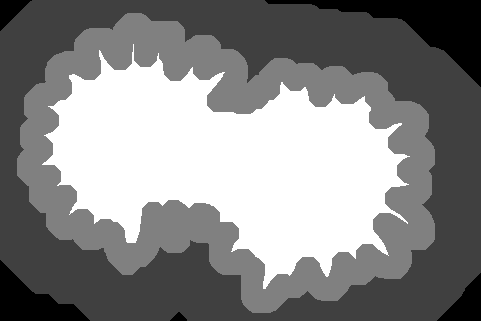}                    } &
\subfloat{\includegraphics[width=2.8cm,height=2.3cm,keepaspectratio]{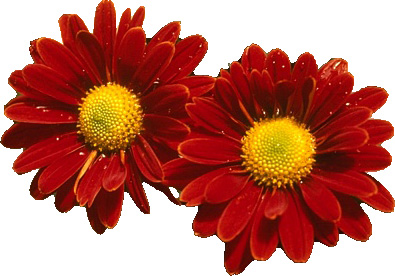}                    } &
\subfloat{\includegraphics[width=2.8cm,height=2.3cm,keepaspectratio]{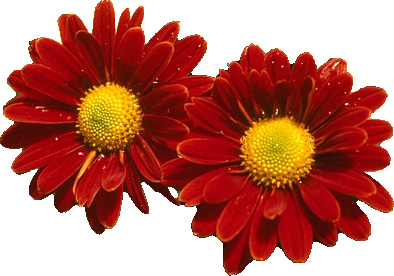}                    } \\
\subfloat{\includegraphics[width=2.8cm,height=4cm,keepaspectratio]{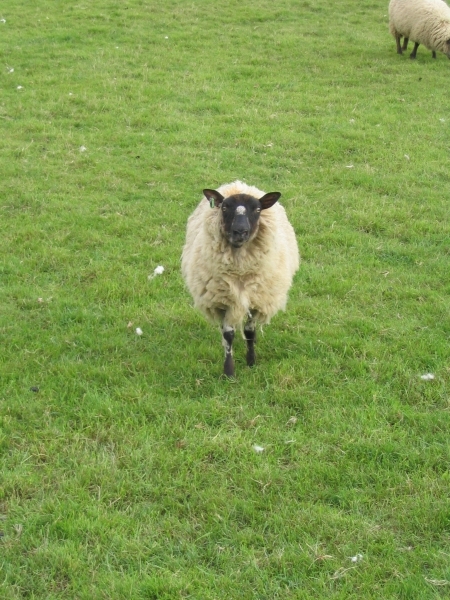}                    } &
\subfloat{\includegraphics[width=2.8cm,height=4cm,keepaspectratio]{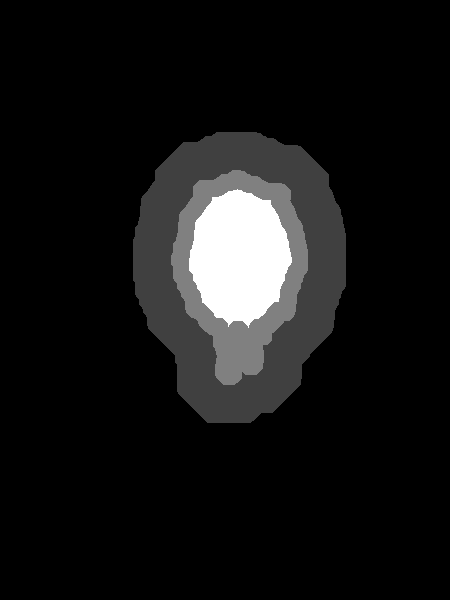}                    } &
\subfloat{\includegraphics[width=2.8cm,height=4cm,keepaspectratio]{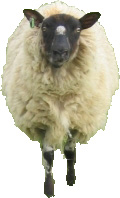}                    } &
\subfloat{\includegraphics[width=2.8cm,height=4cm,keepaspectratio]{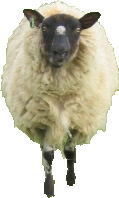}                    } \\
(a) & (b) & (c) & (d) \\
\end{tabular}
\caption{The proposed method applied to the Microsoft GrabCut dataset: (a) input images, (b) ``trimaps'' providing seed regions, (c) close-up foreground segmentation results with default parameters, (d) close-up foreground segmentation results with optimized $k$.}
\label{fig:MicrosoftGrabcut}
\end{figure}

Figure \ref{fig:MicrosoftGrabcut} shows some examples of images from the Microsoft GrabCut dataset and the corresponding segmentation results. The first column shows the input images. The second column show ``trimaps'' providing seed regions. Black (0) represents the background, ignored by the algorithm; dark gray (64) is the labeled background; light gray (128) is the unlabeled region, which labels are estimated by the proposed method; and white (255) is the labeled foreground, which generates the foreground class particles. The error rates in Table \ref{tab:GrabCutComparison} are computed as the ratio of the number of incorrectly classified pixels to the total amount of unlabeled pixels. Third and fourth columns show the segmentation results obtained by the proposed method with its default parameters and with $k$ optimized for each image, respectively.

\subsection{Execution Times}

The algorithm was implemented in MATLAB. The loops in both stages were implemented in C (MEX function). It took an average of $439$ milliseconds to segment each image from the Microsoft GrabCut dataset on a computer with an Intel Core i7 4790K CPU and 32GB of RAM.

\citet{Wang2018b} presents a comparison of the average running times of $7$ representative interactive image segmentation techniques on all $20$ test images of size $321 \times 481$ in the Microsoft GrabCut dataset. They also used an Intel i7 CPU and MATLAB implementations in their tests. Therefore, the same test was applied to the proposed method and the results are shown in Table \ref{tab:GrabcutComparisonTime}. The proposed method was faster than all the other tested methods.

\begin{table}
\caption{Comparison of the average running time obtained on all 20 images of size $321 \times 481$ in the Microsoft GrabCut dataset \citep{Rother2004} by the proposed method and other interactive image segmentation methods, using the original \emph{trimaps}. The times for the other methods were reported by \citet{Wang2018b}.}
\centering
\begin{tabular}{lc}
\hline
{\bf Method} & {\bf Time (s)} \\
\hline
Nonparametric Higher-Order \citep{Kim2010} &       11.0 \\

Multi-layer Graph Constraints \citep{Wang2016} &        5.4 \\

Sub-Markov Random Walk \citep{Dong2016} &        5.1 \\

Diffusive Likelihood \citep{Wang2018b} &        3.4 \\

Laplacian Coordinates \citep{Casaca2014} &        3.2 \\

Grady's random walker \citep{Grady2006} &        0.8 \\

GrabCut \citep{Rother2004} &        0.7 \\

{\bf Proposed Method (default parameters)} &  {\bf 0.3} \\
\hline
\end{tabular}
\label{tab:GrabcutComparisonTime}
\end{table}

\subsection{Parameter Analysis}

\begin{figure}
\centering
\setlength\tabcolsep{1pt}
\begin{tabular}{cc}
\subfloat[]{\includegraphics[height=3.8cm]{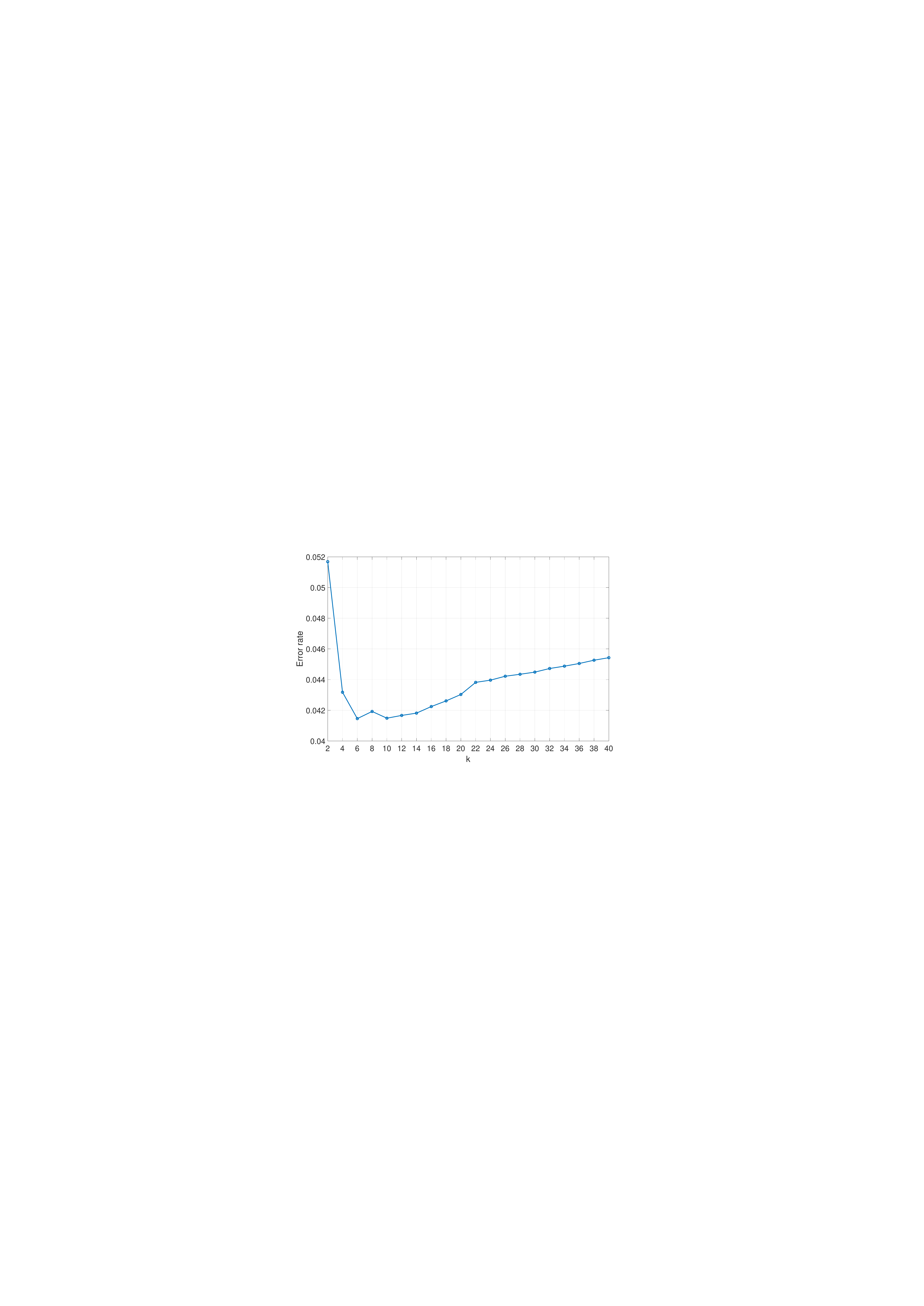}} &
\subfloat[]{\includegraphics[height=3.8cm]{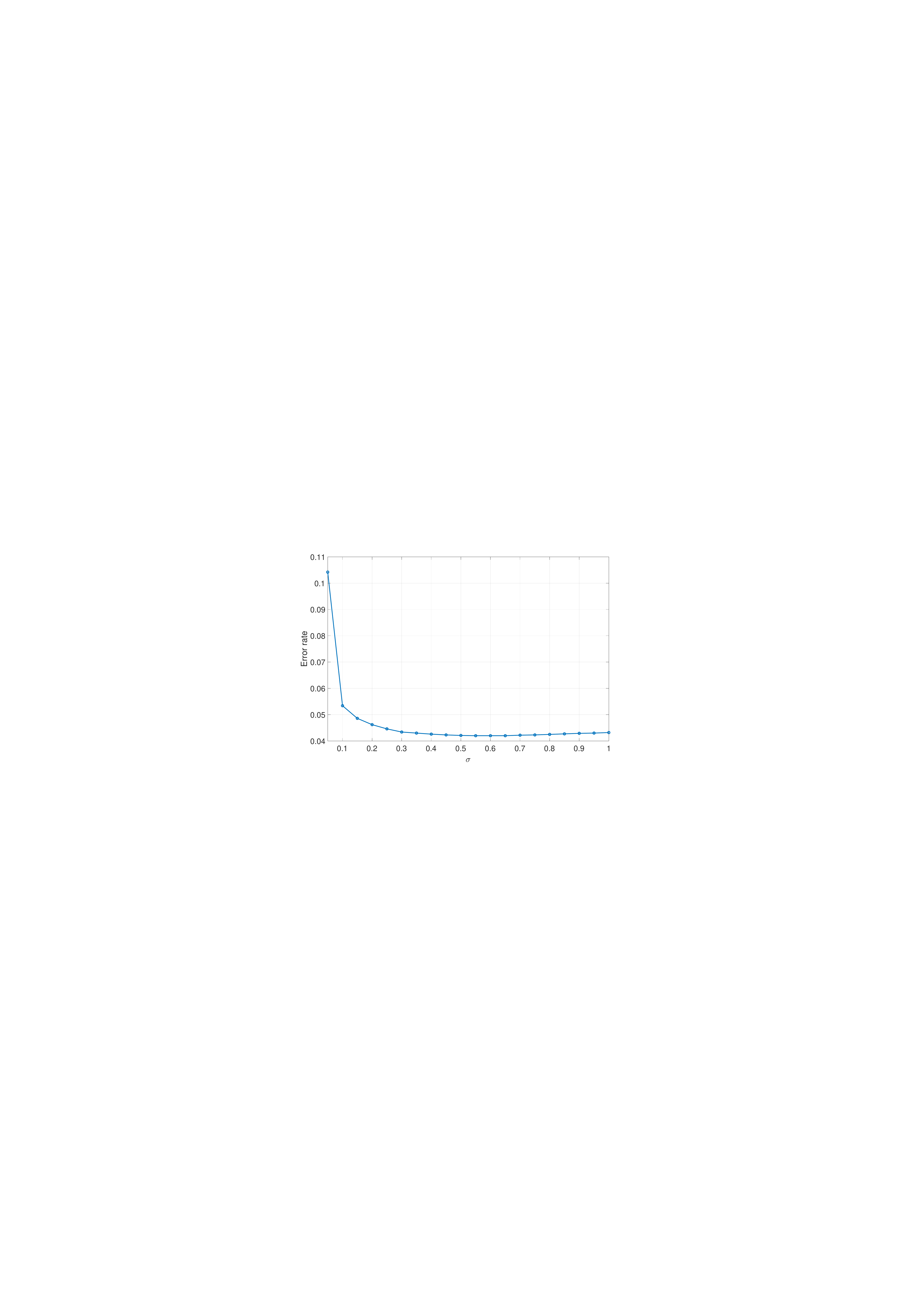}} \\
\subfloat[]{\includegraphics[height=3.8cm]{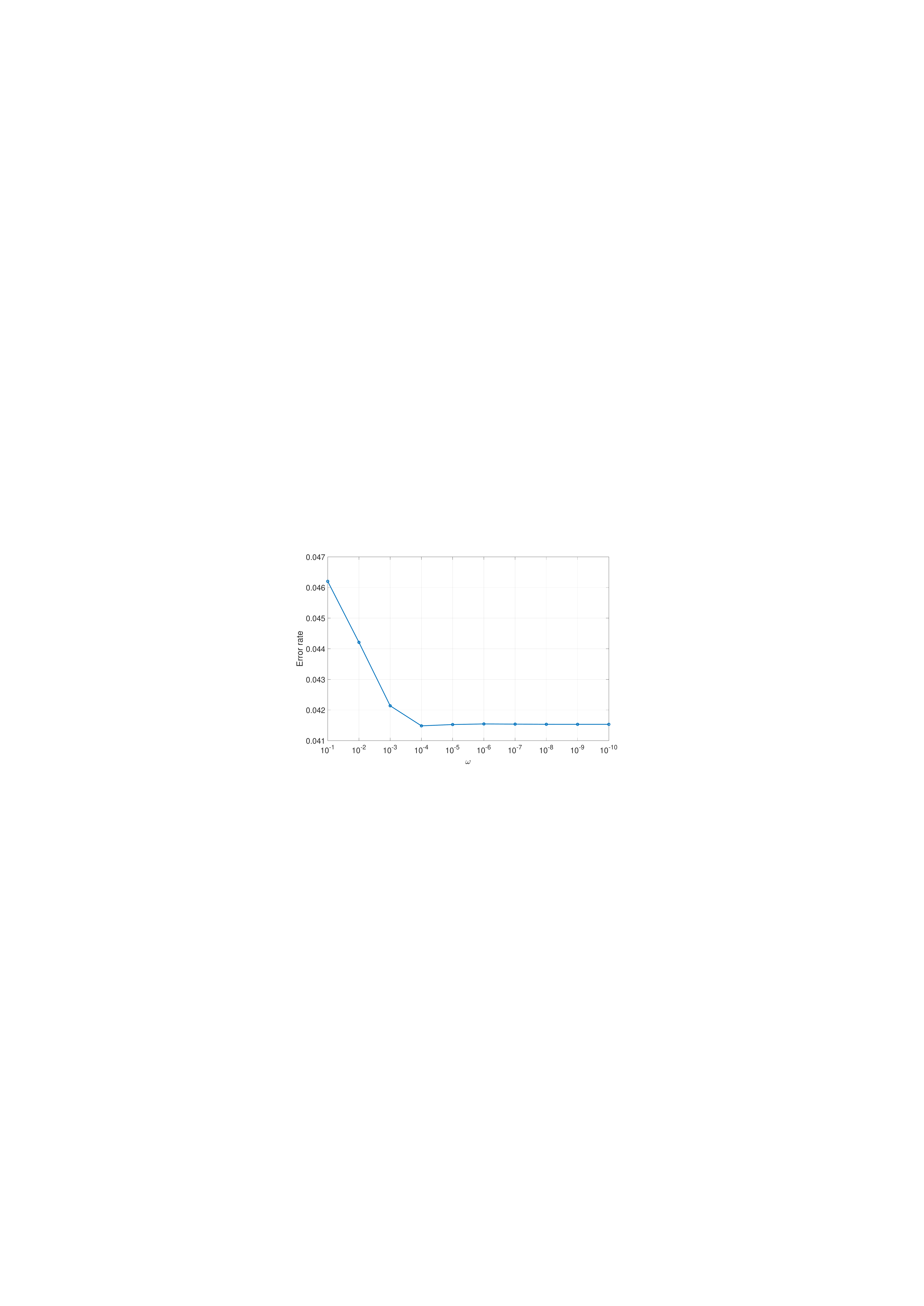}} &
\subfloat[]{\includegraphics[height=3.8cm]{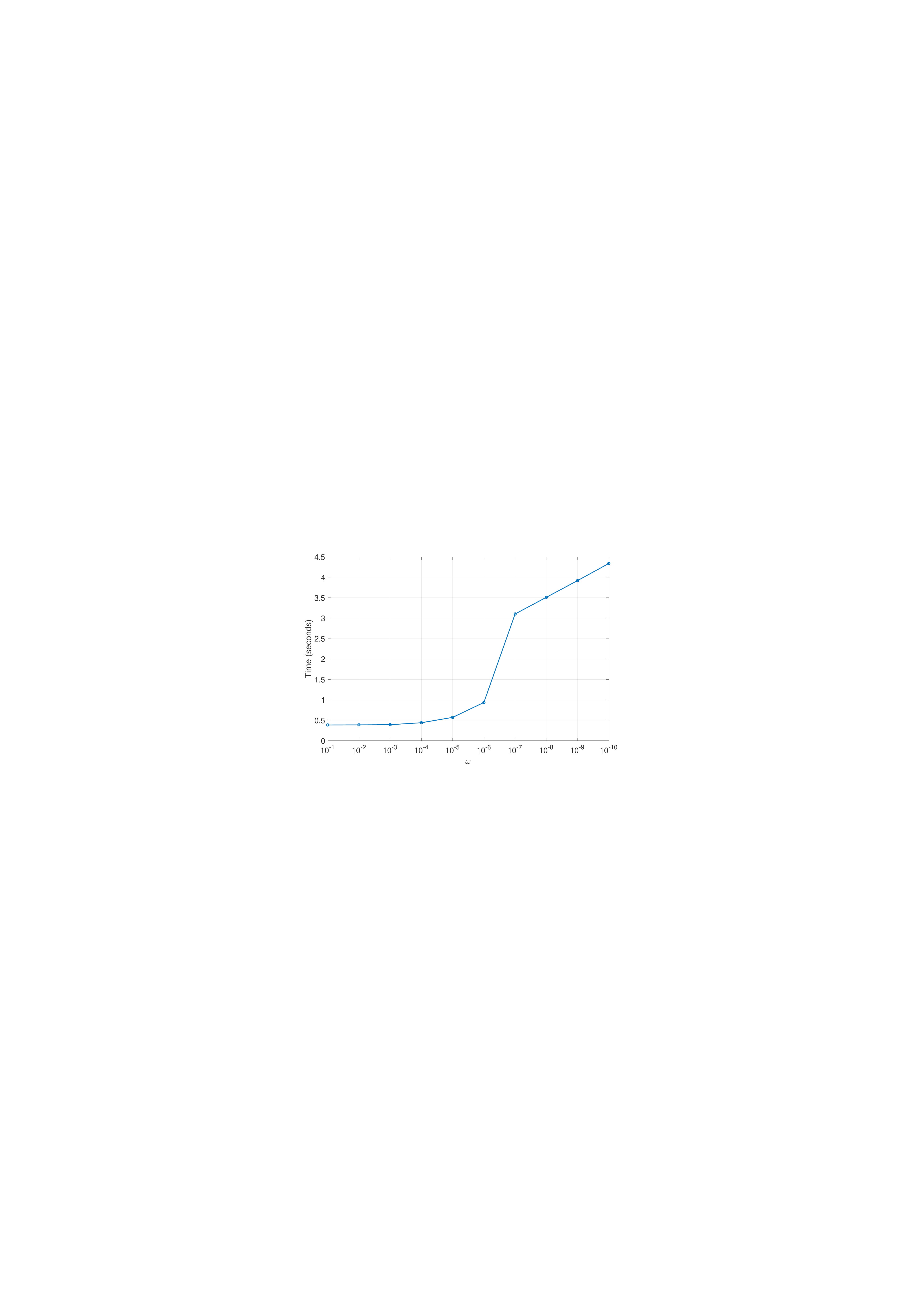}}
\end{tabular}
\caption{The proposed method applied to the Microsoft GrabCut dataset with the default parameters except for the parameter under analysis: (a) error rates for $k=2$ to $k=40$, (b) error rates for $\sigma=0.05$ to $\sigma=1.00$, (c) error rates for $\omega=10^{-1}$ to $\omega=10^{-10}$ , (d) execution times for $\omega=10^{-1}$ to $\omega=10^{-10}$ (average of $100$ realizations).}
\label{fig:ParameterAnalysis}
\end{figure}

The proposed method sensitivity to parameter values is analyzed using the Microsoft GrabCut dataset. In all scenarios, the $50$ images of the dataset are segmented with the default parameters, except for the parameter under analysis. Figure \ref{fig:ParameterAnalysis}(a) shows the error rates when $k=\{2, 4, \dots, 40\}$. Figure \ref{fig:ParameterAnalysis}(b) shows the error rates when $\sigma=\{0.05, 0.10, \dots, 1.00\}$. Figures \ref{fig:ParameterAnalysis}(c) and \ref{fig:ParameterAnalysis}(d) shows the error rates and execution times when $\omega=\{10^{-1}, 10^{-2}, \dots, 10^{-10}\}$.

By analyzing those graphics, one can notice that $k=8$ and $k=10$ produced the best results in the $k$ parameter analysis. $\sigma$ has low sensitivity and has its best range around $\sigma=0.4$ to $\sigma=0.7$. Finally, $\omega$ has decreasing error rates as it lowers down to $\omega=10^{-4}$, and then it stabilizes. However, since this parameter is directly related to the stop criterion, the execution times are higher as $\omega$ decreases. $\omega=10^{-4}$ offers a good trade-off between execution time and error rates.

\subsection{Seed Sensitivity Analysis}

\begin{figure}
\centering
\setlength\tabcolsep{1pt}
\begin{tabular}{cc}
\subfloat[]{\includegraphics[height=3.8cm]{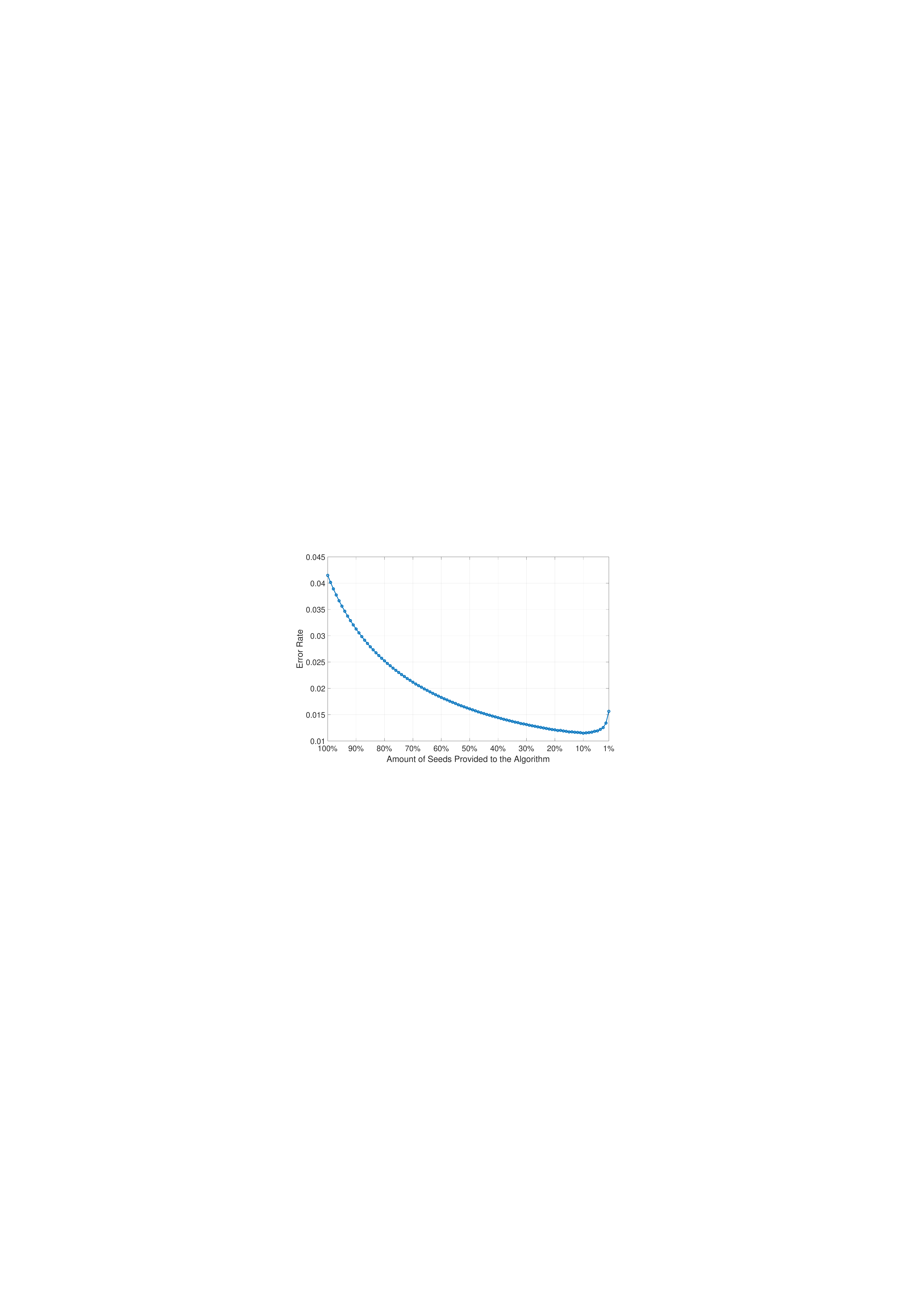}} &
\subfloat[]{\includegraphics[height=3.8cm]{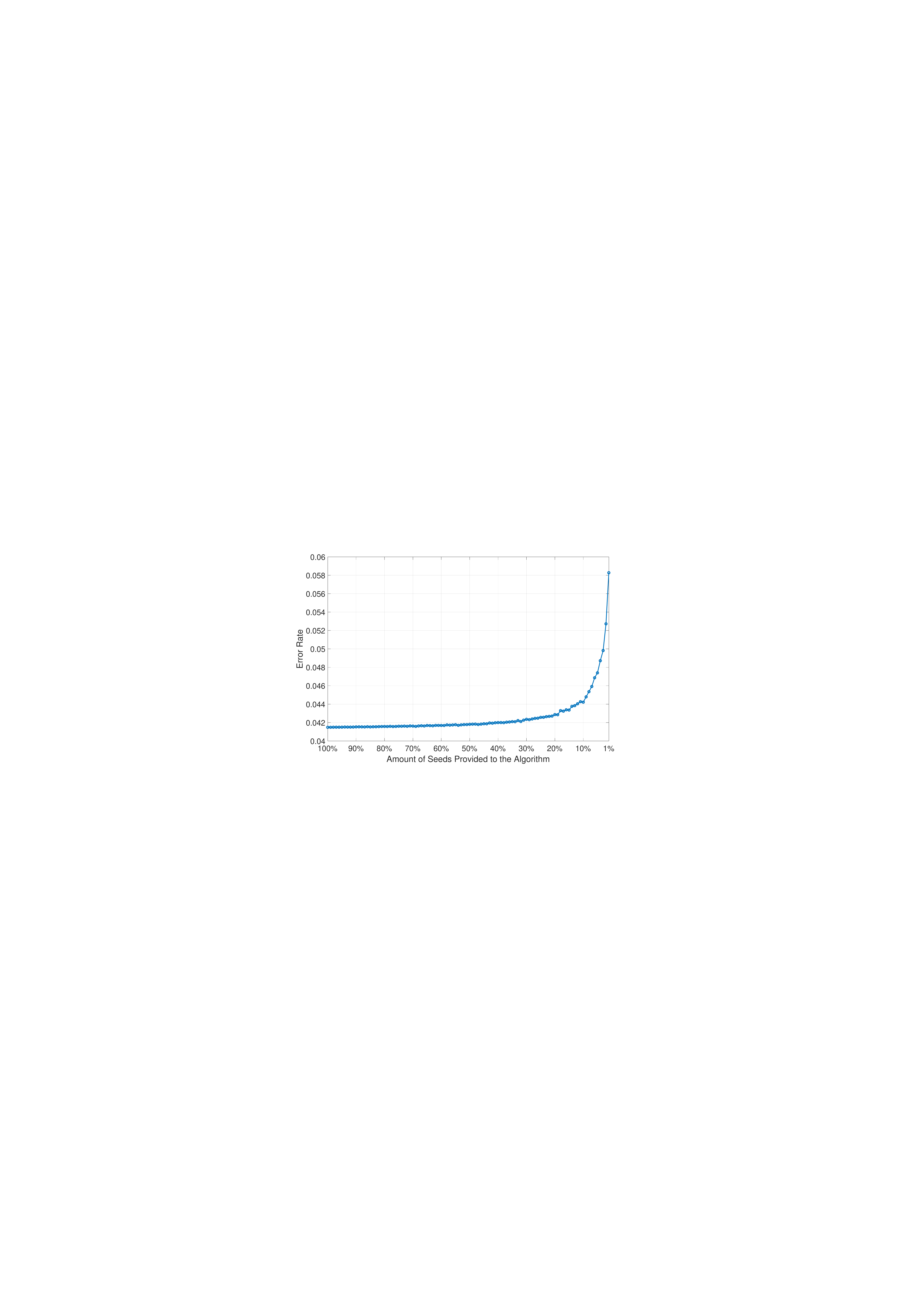}} \\
\end{tabular}
\caption{The proposed method with its default parameters applied to the Microsoft GrabCut dataset with a subset of the original seeds. Each point is an average of $20$ realizations with different random seeds selected: (a) all unlabeled pixels are computed in the error rate, (b) unlabeled pixels that were originally a seed pixel are not computed in the error rate.}
\label{fig:SeedReduction}
\end{figure}

The original ``trimaps'' from the Microsoft GrabCut dataset provides a large number of seeds for iterative image segmentation methods. However, the proposed method does not need all those seeds to provide reasonable segmentation results. Therefore, an experiment was set in which each ``trimap'' from the dataset had each of its seeds randomly erased with a probability $p$, so the changed pixels would appear unlabeled to the method. By varying $p$ from $0$ to $0.99$, it is possible to generate ``trimaps'' with roughly 100\% to 1\% of the original seeds, respectively. So $2,000$ ``trimaps'' were generated for each image, $20$ of them for each of the $100$ configurations of $p = \{ 0.00, 0.01, \dots, 0.99\}$. The proposed method was applied to all of them. The mean error rates on each configuration are presented in Figure \ref{fig:SeedReduction}(a).

Notice that while the error rates decrease as the number of seeds decreases in Figure \ref{fig:SeedReduction}(a), that does not necessarily mean that the segmentation results are better, because with fewer seeds, there are more unlabeled pixels and each pixel mislabeled by the algorithm has less impact on the error rate. Thus, Figure \ref{fig:SeedReduction}(b) shows the error rates on each configuration, but excluding the pixels which were seeds in the original ``trimaps'' from the error rate computation. These results showed that the number of seeds may be greatly reduced without much impact in the error rates.

\subsection{Microsoft GrabCut Dataset with ``Scribbles''}

\citet{Andrade2015} presents an objective and empirical evaluation method for seed-based interactive segmentation algorithms. They have extended the Microsoft GrabCut dataset by incorporating two sets of ``scribbles'' for each of the $50$ images\footnote{Available at \url{https://github.com/flandrade/dataset-interactive-algorithms}}.

The first set of ``scribbles'' employ four strokes per image, three on the background and one small area on the foreground object. The second set of ``scribbles'' indicate and mark in more detail the foreground region. The two sets reflect two different degrees of user effort.

The proposed method was applied to both sets of ``scribbles''. Table \ref{tab:DIA} presents a comparison of the average error rates obtained by the proposed method and other interactive image segmentation methods. The proposed method was first applied to the whole dataset with its default parameters ($k=10$, $\mathbf{\lambda} = \lambda_1$, $\omega=10^{-4}$). Later, the parameter $k$ was optimized for each image. In both scenarios, and for both sets of ``scribbles'', the proposed method outperformed the other $8$ methods even with its default parameters.

\begin{table}
\caption{Comparison of the average error rates obtained on the GrabCut dataset \citep{Rother2004} by the proposed method and other interactive image segmentation methods, using  the sets of ``scribbles'' from \citet{Andrade2015} The error rates for the other methods were reported by \citet{Bampis2017}.}.
\centering
\resizebox{\textwidth}{!}{%
\begin{tabular}{lcc}
\hline
           & \multicolumn{ 2}{c}{{\bf Error}} \\

{\bf Method} & {\bf $S_1$} & {\bf $S_2$} \\
\hline
Random Walks with Restart \citep{Kim2008} &     6.65\% &     6.44\% \\

Lazy Randow Walks \citep{Shen2014} &     6.42\% &     6.12\% \\

Normalized Sub-Markov Random Walk \citep{Bampis2017} &     6.07\% &     5.81\% \\

Sub-Markov Random Walk \citep{Dong2016} &     6.07\% &     5.81\% \\

Grady's random walker \citep{Grady2006} &     5.58\% &     2.91\% \\

Laplacian Coordinates \citep{Casaca2014} &     5.37\% &     3.75\% \\

Normalized Lazy Random Walker \citep{Bampis2017} &     4.80\% &     2.49\% \\

Normalized Random Walker \citep{Bampis2017} &     4.77\% &     2.48\% \\

{\bf Proposed Method (default parameters)} & {\bf 3.68\%} & {\bf 1.60\%} \\

{\bf Proposed Method (optimized $k$)} & {\bf 2.28\%} & {\bf 1.21\%} \\
\hline
\end{tabular}
}
\label{tab:DIA}
\end{table}

\section{Conclusions}
\label{sec:Conclusions}

In this paper, a graph-based interactive image segmentation method is proposed. Seeds are provided by the user in form of ``scribbles'', loosely traced over the objects of interest and the background. The method takes advantage of complex networks properties to spread labels quickly, with low time and storage complexity. It can be applied to multi-class problems at no extra cost.

Despite its simplicity, the method can achieve classification accuracy comparable to those achieved by some state-of-the-art methods when applied to the Microsoft GrabCut dataset, with their original \emph{trimaps} used as user input, which is commonly used to evaluate and compare interactive image segmentation methods. It is also the fastest method when compared to other $7$ methods, including some classic and some newer state-of-the-art approaches. Moreover, it achieved the best results when the user input is composed only by a few ``scribbles'', outperforming $8$ other recent approaches.

Though the proposed method has some parameters which can be fine-tuned to achieve better results, usually only $k$ has a significant impact on the classification accuracy. The default parameters may be used when the time is restricted. The user may also fine-tune parameters while adding more ``scribbles'' if he/she is not satisfied with the current segmentation results.

The method may also be extended by introducing edge-finding components or edge related features to decrease error rates further, and to handle more challenging segmentation tasks.

\section*{Acknowledgment}

The author would like to thank the S\~{a}o Paulo State Research Foundation - FAPESP [grant 2016/05669-4].

\bibliographystyle{elsarticle-harv}
\bibliography{ESWA}

\end{document}